\documentclass[preprint,3p,times]{elsarticle} 

\usepackage{natbib}
\usepackage{mathrsfs}

\usepackage{times}
\usepackage{graphicx}
\usepackage{footnote}
\usepackage{array}
\usepackage{comment}
\usepackage{graphicx}
\usepackage{placeins}
\usepackage{float}
\usepackage{subfigure}
\usepackage[table,xcdraw]{xcolor}
\usepackage{url}
\usepackage{graphicx}
\usepackage{array}
\usepackage[inline]{enumitem}
\usepackage{tabularx}
\usepackage{caption}
\usepackage{float}
\usepackage{tabu}
\usepackage{multicol}
\usepackage{multirow}
\usepackage{longtable}
\usepackage{flushend}
\usepackage{adjustbox}
\usepackage{anyfontsize}
\usepackage{footnote}
\makesavenoteenv{tabular}
\makesavenoteenv{table}
\usepackage{lipsum}
\usepackage{courier}
\usepackage{adjustbox}

\usepackage{booktabs} 
\usepackage{csquotes}
\newcommand{\dq}[1]{\enquote{#1}}

\usepackage{hyperref}

\usepackage{amssymb} 
\usepackage{enumitem}
\usepackage{subfigure}
\usepackage{float}



\journal{Arxiv}

\begin{document}%

\begin{frontmatter}

	\title{Word Embeddings for Sentiment Analysis: A Comprehensive Empirical Survey}
	
	\author{Erion \c Cano\fnref{cor1}}
	\ead{erion.cano@polito.it} 	
	
	\author{Maurizio Morisio}
	\ead{maurizio.morisio@polito.it}
	
	\fntext[cor1]{Corresponding Author}
	
	\address{Department of Control and Computer Engineering, Politecnico di Torino, Corso 
		Duca degli Abruzzi, 24 - 10129 Torino}

\begin{abstract}
This work investigates the role of factors like training method, training corpus size and thematic relevance of texts in the performance of word embedding features on sentiment analysis of tweets, song lyrics, movie reviews and item reviews. We also explore specific training or post-processing methods that can be used to enhance the performance of word embeddings in certain tasks or domains. Our empirical observations indicate that models trained with multi-thematic texts that are large and rich in vocabulary are the best in answering syntactic and semantic word analogy questions. We further observe that influence of thematic relevance is stronger on movie and phone reviews, but weaker on tweets and lyrics. These two later domains are more sensitive to corpus size and training method, with Glove outperforming Word2vec. \dq{Injecting} extra intelligence from lexicons or generating sentiment specific word embeddings are two prominent alternatives for increasing performance of word embedding features.
\end{abstract}


\end{frontmatter}

\section{Introduction}
\label{sec:introduction}
Until end of the 90s, the dominant natural language representation models were discrete vector-space representations that relied on big and sparse matrices. Bag-of-Words (BOW) representation and feature extraction methods such as count vectorizer or Tf-idf were particularly popular. A peculiar insight was proposed in 2003 at \cite{Bengio:2003:NPL:944919.944966}, substituting sparse n-gram models with word embeddings which are continuous and denser representations of words. They are obtained by using feed-forward or more advanced neural network architectures to analyze co-occurrences of words that appear together within a context window. 
One of the first popular word embedding models is C\&W (for Collobert and Weston)  which was presented in \cite{Collobert:2008:UAN:1390156.1390177}. They utilized a Convolution Neural Network (CNN) architecture to generate word vectors and use them for several well-known tasks like named entity recognition, part-of-speech tagging, semantic role-labeling etc. Authors experimented with a Wikipedia corpus and reported significant improvements in each NLP task, with no hand-crafted features used. 
\par 
Higher quality and easier to train methods like Continuous Bag-of-Words (CBOW) and Skip-Gram, make use of shallow neural network architectures to either predict a word given a context window (the former), or predict the context window for a given word (the later). These methods were first presented in \cite{Mikolov2013} and considerably improved in \cite{DBLP:conf/nips/MikolovSCCD13} with the introduction of negative sampling and sub-sampling of frequent words. 
Later on in \cite{DBLP:conf/nips/MnihK13}, a light log-bilinear learning approach was presented. It simplifies the training process by using Noise-Contrastive Estimation (NCE), a method that discriminates between data distribution samples and noise distribution by means of logistic regression classification. They claim that NCE applied in distributed word representations produces word vectors of the same quality as Word2vec but in about half of the time. 
\par 
One of the most recent and popular word embedding generation methods is Glove presented in \cite{pennington2014glove}. It uses word-word co-occurrence counts in the co-occurrence matrix to train text corpora, making efficient use of global corpus statistics and also preserving the linear substructure of Word2vec method. Authors provide evidence that Glove scales well on larger text corpora and claim that it outruns Word2vec and other similar methods in various tasks like word analogy. Their experiments were conducted using large text corpora that were also released for public use. 
The wide applicability of word embeddings in various applications like machine translation, error correcting systems or sentiment analysis, is a result of their \dq{magical} ability to capture syntactic and semantic regularities on text words. 
As a result, vectors of words that are syntactically or semantically related with each other, reveal same relation in the vector space. It is thus possible to answer word analogy questions of the form \dq{A is to  B as C is to ?} by performing $V_? = V_B - V_A + V_C$ algebraic operation and then finding the word that best matches $V_?$.  
\par 
The scope of this survey work is to examine the quality of Glove and Word2vec word embeddings on word analogy task as well as on four sentiment analysis tasks: sentiment analysis of tweets, song lyrics, movie reviews, and product (phone) reviews. 
Here, the work in \cite{ccano2017} is extended both quantitatively and qualitatively. Much bigger text collections are used for generating word embeddings which are evaluated on more tasks. 
We were particularly interested to know how do factors like training method and corpus thematics or size influence the quality of word vectors that are produced. For this reason, we collected various pretrained (with Glove or Word2vec) and publicly available corpora and also created other ones of different sizes, domains or content ourself.
\par
Based on our experimental results, best models on word analogy tasks are those obtained from text bundles that are multi-thematic, big in size and rich in vocabulary. We also observed that training method and training corpus size do have a strong influence on sentiment analysis of lyrics, a moderate one on tweets and no significant influence on movie and phone reviews. Contrary, thematic relevance of training corpus has no significant influence on song lyrics, a moderate one on tweets and a strong influence on movie and phone reviews. We also consulted relevant research works that try to improve quality of word embeddings by means of post-processing or improved training methods. We assessed two such methods and observed that they do improve the quality of word embeddings at some scale, at least on certain tasks. 
These results might serve as guidelines for researchers or practitioners who work with word embeddings on sentiment analysis or similar text-related tasks. 
Anyone interested in our trained models can download them from \url{http://softeng.polito.it/erion/} to reproduce our experiments or perform any related experiments as well.  
\par 
The rest of this paper is organized as follows: In Section~\ref{experimentalDesign} we present the research questions, show the preprocessing and training steps and describe the corpora we experiment with. In Section~\ref{tasksAndDatasets} we present related state-of-the-art works in the four sentiment analysis tasks and also describe the datasets we used in our experiments. Section~\ref{linguisticRegularity} presents results of the different models on word analogy task. Role of attributes such as training method, corpus size and thematic relevance of texts on model performance is examined on Section~\ref{corpusCharacteristics}.  Section~\ref{improvingQuality} presents several proposed improvements on quality of word embeddings and finally, Section~\ref{discussion} ends our discussion.  
\section{Experimental Design}
\label{experimentalDesign}
\subsection{Research Questions}
\label{researchQuestions}
The purpose of this study is to examine the quality of word embeddings, observe how training method and corpus attributes influence their it, and assess their performance when used as classification features on sentiment analysis tasks. We conducted experiments in four tasks: \emph{(i)} sentiment analysis of tweets, \emph{(ii)} sentiment analysis of song lyrics, \emph{(iii)} sentiment analysis of movie reviews and \emph{(iv)} sentiment analysis of phone reviews. The four research questions we defined are:
\begin{enumerate}
	\itemsep0.7em 
	\item[\hypertarget{rq1}{\textbf{RQ1}}] \emph{How do corpus size, corpus theme and training method influence semantic and syntactic performance of word embeddings on word analogy tasks?}
	\item[\hypertarget{rq2}{\textbf{RQ2}}] \emph{How do corpus size, corpus theme and training method influence performance of word embedding features in the above four tasks?}
	\item[\hypertarget{rq3}{\textbf{RQ3}}] \emph{What post-processing methods have been proposed and how do they affect quality of word embeddings?}
	\item[\hypertarget{rq4}{\textbf{RQ4}}] \emph{What alternative training methods have been proposed and how do they compare with the popular ones?} 
\end{enumerate}
We answer the research questions based on our experimental results as well as by comparing those results with recent and relevant research works. For RQ1 we assess the semantic and syntactic quality of each text bundle using word analogy questions. To answer RQ2 we conducted many experiments with the full set of text corpora as well as several cuts of different sizes for each task. RQ3 and RQ4 are answered in Section~\ref{improvingQuality} where we discuss and assess methods from literature that try to improve the quality of word embeddings.  
\subsection{Text Preprocessing and Training}
\label{textPrep}
Before training the corpora, we applied few preprocessing steps removing html tags (some texts were scraped from the web), removing punctuation symbols and lowercasing everything. We also applied word-level English language filter to remove every word that was not in English. We did not remove stopwords as they are quite frequent and useful when generating word vectors. Same preprocessing steps were also applied to the four experimentation datasets that are described in each subsection of Section~\ref{tasksAndDatasets}. 
The basic text bundles we used, together with some of their attributes are listed in Table~\ref{table:corporaProperties}. The first six are pretrained with either Glove or Word2vec. We trained the other six using both methods and 18 basic models were obtained in total. 
Throughout this paper, we use the term \dq{tokens} to denote the total number of words inside a text collection that repeat themselves in it. The total number of unique words (vocabulary size) of the collection is simply called words. In \dq{Size} column we show the number of tokens in training corpora after preprocessing but before applying the training methods. \dq{Dimensions} on the other hand, presents the number of float values each word of the resulting model is associated with. For consistency with the pretrained models, we trained our corpora with 300 dimensions. As reported in \cite{pouransari2014deep} where authors conduct similar experiments, 100 - 300 dimensions are usually enough and higher dimension vectors do not provide any performance gain in most of the applications. 
We also used $min\_count=5$ as the minimal frequency of words to be kept in each model. For this reason, the values reported in \dq{Vocabulary} column represent the number of words appearing 5 or more times in the training corpus. We chose a context window size of 10 tokens when training each model. In the following subsection, we provide some more details about the origin and content of the text bundles that we created and used.
\begin{table}[ht] 
	\centering   
	\caption{Properties of basic word embedding models}  
	\small 
	\setlength\tabcolsep{3.2pt}    
	\begin{tabular}
		{l | c c c c c}  
		\toprule
		\textbf{Corpus Name} & \textbf{Training} & \textbf{Size} & \textbf{Vocabulary} & \textbf{Dimensions} & \textbf{Download}  	\\ [0.1ex] 
		\midrule   
		WikiGigaword & Glove & 6B & 400000  & 300 float32 & \href{http://nlp.stanford.edu/projects/glove/}{link}			  \\ [0.2ex] 
		CommCrawl840 & Glove & 840B & 2.2M & 300 float32 & \href{http://nlp.stanford.edu/projects/glove/}{link}			  \\ [0.2ex] 
		CommCrawl42 & Glove & 42B & 1.9M  & 300 float32 & \href{http://nlp.stanford.edu/projects/glove/}{link}			  \\ [0.2ex] 
		WikiDependency & W2v & 1B & 174000  & 300 float32 &  
		\href{https://levyomer.wordpress.com/2014/04/25/dependency-based-word-embeddings/}{link} \\ [0.2ex] 
		GoogleNews & W2v & 100B & 3M  & 300 float32 & \href{https://code.google.com/archive/p/word2vec/}{link}			  \\  [0.2ex] 
		TwitterTweets  & Glove & 27B & 1.2M & 200 float32 & \href{http://nlp.stanford.edu/projects/glove/}{link}			  \\ [0.2ex] 
		MischBundle & Both & 63B & 1.7M & 300 float32 & \href{http://softeng.polito.it/erion/}{link}			  \\ [0.2ex] 
		TweetsBundle & Both & 500M & 96055 & 300 float32 & \href{http://softeng.polito.it/erion/}{link}			  \\ [0.2ex] 
		MoodyCorpus & Both & 24B & 71590 & 300 float32 & \href{http://softeng.polito.it/erion/}{link}			  \\ [0.2ex] 
		ItemReviews & Both & 2.5B & 82682 & 300 float32 & \href{https://cs.textdata.html}{link}			  \\
		NewsBundle & Both & 1.3B & 171000 & 300 float32 & \href{http://nlp.stanford.edu/projects/glove/}{link}			  \\ [0.2ex] 
		Text8Corpus & Both & 17M & 71290 & 300 float32 & \href{https://cs.fit.edu/\%7Emmahoney/compression/textdata.html}{link}			  \\
		\bottomrule
	\end{tabular} 
	\label{table:corporaProperties}
\end{table}
\subsection{Contents of Text Sets}
\label{bundleContents}
The first text collection listed in Table~\ref{table:corporaProperties} is Wikipedia Gigaword, a combination of Wikipedia 2014 dump (1.6 billion tokens) and Gigaword 5 (4.3 billion tokens) with about 6 billion tokens in total. It was created by authors of \cite{pennington2014glove} to evaluate Glove performance. 
Wikipedia Dependency is a collection of 1 billion tokens from Wikipedia. The method used for training it is a modified version of Word2vec described in \cite{DBLP:conf/acl/LevyG14}. Authors experiment with different syntactic contexts of text, rather than using the linear bag-of-words context. 
They report that the extra contexts they introduce produce different embeddings resulting in more functional similarities between words. 
\emph{GoogleNews} is one of the biggest and richest text sets around, consisting of 100 billion tokens and 3 million words and phrases. As explained in \cite{Mikolov2013}, they added word phrases to the basic texts to further increase the richness of the dataset. The version we use here was trained using CBOW Word2vec with negative sampling, windows size 5 and 300 dimensions. Reduced versions of this corpus were also used in \cite{DBLP:conf/nips/MikolovSCCD13} for validating the efficiency and training complexity of Word2vec.
\emph{CommonCrawl840} is even bigger than \emph{GoogleNews}. It is a huge corpus of 840 billion tokens and 2.2 million word vectors, also used at \cite{DBLP:conf/nips/MikolovSCCD13} to evaluate the scalability of Glove. This bundle contains data of Common Crawl (\url{http://commoncrawl.org/}), a nonprofit organization that builds and maintains free and public text sets by crawling the Web. \emph{CommonCrawl42} is a highly reduced version easier and faster to work with. It is made up of 42 billion tokens and a vocabulary of 1.9 million words. Both \emph{CommonCrawl840} and \emph{CommonCrawl42} were trained with Glove producing word vectors of 300 dimensions. 
The last pretrained corpus we used is \emph{TwitterTweets}, a collection of tweets also trained with Glove. It consists of 2 billion tweets from which 27 billion tokens with 1.2 million unique words were extracted. From the four available models, we picked the one with 200 (highest) dimension vectors.   
\par
Besides the above six, we also used six more corpora that we built and trained ourself. \emph{MischBundle} is the biggest, consisting of 63 billion tokens and a vocabulary of 1.7 million words. We made sure to incorporate a high variety of text types and sources inside. 
The biggest component was a text set of 500-600 million comments filtered from Self-Annotated Reddit Corpus (SARC). It was extracted from Reddit social site and used for sarcasm detection in \cite{SARC}. 
Another important component we used is the reduced USENET corpus created by authors of \cite{shaoul2013reduced}. It consists of public USENET postings collected between Oct 2005 and Jan 2011 from 47,860 English groups. 
We also mixed in the English version of European Parliament Proceedings Parallel Corpus, a text collection extracted from proceedings of European Parliament between 1996 - 2011. It was created to generate sentence alignments of texts in different languages for machine translation. Further specifications can be found in \cite{koehn2005epc}.
\emph{TwitterBundle} is a collection of Twitter tweets consisting of 500 million tokens and a vocabulary of 96055 words. 
The most important component we used to create it is a dataset of about 9 million public tweets from 120,000 Twitter users. It was collected by authors of \cite{cheng2010you} from September 2009 to January 2010 and used to explore geolocation data. 
The other collection of tweets we used was released by Sentiment140 (\url{http://help.sentiment140.com/}) project, a service that provides Twitter opinions about brands, products or topics. 
\emph{MoodyCorpus} is the collection of many song texts that we scraped from different music websites from March to June 2017. There are 24 billion tokens in the source and  71590 unique words in the generated model. Same as in the other cases, we kept English texts only, dropping out songs in other languages. To have a diversified collection with rich terminology, we tried to pick up songs of different genres and time periods.  
\par
For movie and phone review experiments we created \emph{ItemReviews}, a text set of 2.5 billion tokens and 82682 words. 
We mixed in a set of 142 million Amazon product reviews of about 24 item categories such as movies, books or electronic devices, spanning from May 1996 to July 2014. Authors used it in \cite{mcauley2015image} to build a recommender system of products modeling item relationships as seen by people. 
Another source we used is the collection of movie critics we extracted from Movie\$ Data corpus, a set of metadata, financial information and critic reviews about movies. Authors used that dataset in \cite{smith461movie} to predict movies' revenues in their opening weekends. 
News articles and headlines are very good sources of texts for topic modeling experiments. For this reason, we built \emph{NewsBundle}, a text body of 1.3 billion tokens appearing in online news and headlines. The resulting model has 171000 words and vectors.  
To construct it, among other sources we made use of AG's corpus of news articles (\url{https://www.di.unipi.it/gulli/AG_corpus_of_news_articles.html}). 
It is a collection of over 1 million news articles gathered from more than 2000 news sources by ComeToMyHead, an academic search engine. 
We also utilized various news collections uncopyrighted and freely distributed in hosting archives (\url{https://webhose.io/datasets}). To better observe the role of corpus size in the quality of generated embeddings, we also trained and used \emph{Text8}, a comparably tiny collection consisting of 17 million tokens. It contains the first $10^9$ bytes of English Wikipedia dump collected in March 2006 and is often used as a trial corpus for illustrating word embeddings. The resulting model has 71290 words and vectors.    
\section{Sentiment Analysis Tasks and Datasets}
\label{tasksAndDatasets}
\subsection{Sentiment Analysis of Tweets} 
\label{tweetSent}
Sentiment analysis of tweets has become popular due to the increasing use and popularity of Twitter microblogging website. The objective here is to identify the sentiment polarity of the informal tweets and then use it for large-scale opinion mining about various products, companies, events etc. Several lexicon-based studies like \cite{Ding:2008:HLA:1341531.1341561} or \cite{taboada2011lexicon} use dictionaries of affect words with their associated polarity for computing the overall sentiment polarity of each message. Machine learning or more recently deep learning methods, on the other hand, utilize abundant sets of tweets to evaluate unsupervised or supervised models they construct. In \cite{hu2013unsupervised} they explore the use of emoticons and other emotion signals in an unsupervised framework for sentiment analysis. Authors demonstrate the effectiveness of their framework by comparing it with state-of-the-art techniques and also illustrate the behavior of different emotion signals in sentiment analysis. 
Another relevant study is \cite{Severyn:2015:TSA:2766462.2767830} where authors first use a CNN to refine the previously trained word embeddings and then train that same CNN on a Semeval-2015 (\url{http://alt.qcri.org/semeval2015/}) dataset of tweets to get the final classifier. Their model achieves an F1 score of 84.79 \% on phrase-level (subtask A) and 64.59 \% on message-level (subtask B) classification. 
There are also studies like \cite{DBLP:journals/corr/MohammadKZ13} that focus entirely on feature engineering and utilize off-the-shelf methods for classification. The authors mixed together a high variety of hand-crafted features like word n-grams, character n-grams or hashtags together with emoticons and lexicon features. They used Semeval-2013 (\url{http://www.cs.york.ac.uk/semeval-2013/task2}) data to train a linear kernel SVM and were ranked in the first place of Semeval-2013 competition, achieving an F1 score of 68.46 \% on message-level task and 89.10 \% on term-level task. 
Authors in \cite{tang2014coooolll} present Coooolll, a supervised deep learning system for Twitter sentiment classification that mixes together state-of-the-art hand-crafted text features with word embedding features. They concatenate N-dimensional embedding features with K features from \cite{DBLP:journals/corr/MohammadKZ13} and feed everything to the neural network. Authors evaluate Coooolll in Semeval-2014 Twitter dataset and push the F1 score up to 87.61\% which is more than 10\% higher compared to pure word2vec accuracy. 
The dataset we used for Twitter sentiment analysis is a combination of $train$ and $dev$ parts of Semeval-2013 Task2 together with test sets of Semeval-2014 Task9. We removed the duplicate tweets and also those that were not available. For consistency with the other experiments, we also removed the \dq{neutral} tweets, keeping only the remaining 12,300 positive and 6711 negative ones. Finally, we equally balanced the 2 categories reaching a total of 13,400 tweets in the final set. 
\subsection{Sentiment Analysis of Song Lyrics}
\label{lyricSent}
Music mood recognition or music sentiment analysis is about utilizing machine learning, data mining and other techniques in combination with various feature types to classify songs in two or more categories of emotions. Combinations of several feature types such as audio, lyrics or even social community tags are involved in the process. 
Signal processing methods of audio were the first to be used. 
Studies using lyrics are more recent and fall into two categories: lexicon-based and corpus-based \cite{Tang:2015:DLS:2854473.2854476}. The former utilize dictionaries of sentiment words and polarity norms whereas the later construct models based on annotated datasets and machine learning algorithms. 
With the recent boost of social networks and the huge amount of user data generated every day, social community tags are becoming a very powerful means for extracting knowledge (the so-called collective intelligence). There are however problems like polysemy or lack of a common tag vocabulary that have hindered the use of social tags. Various studies have tried to facilitate and promote their use. In \cite{citeulike:6285088} for example, authors use \emph{Last.fm} songs and tags and also LSA to create a semantic space of music mood categories. They reduce the space into a categorical 2D model with 4 music mood classes, highly compliant with the popular model of Russell \cite{citeulike:630522}. 
Viability and effectiveness of social tags and other crowdsourcing alternatives for creating labeled datasets is also discussed in \cite{pub2677905}. In music domain, we use MoodyLyricsPN, a datasets of English song texts constructed utilizing a lexicon-based approach. It consists of 2595 song lyrics labeled with one of the four categories of Russell's emotional model \cite{Russell1980}. It was constructed following a systematic process based on crowdsoured \emph{Last.fm} user tags \cite{aiap17}. 
\subsection{Sentiment Analysis of Movie and Phone and Reviews} 
\label{itemMovieSent}
Item review sentiment analysis is about training intelligent models that are able to predict user satisfaction level from various kinds of products, based on the textual and other descriptions of those products provided by previous users. This task has a high commercial interest as it is a basic component of advertising recommender systems. The later are intelligent engines that suggest relevant items to users based on item characteristics and prior user preferences. Our focus here is on movie reviews which are very popular and receive a lot of feedback (in terms of text reviews, stars etc.) on the Internet. We also experiment with reviews of phones that are sold on Amazon.  
Foundational works in movie review polarity analysis are \cite{Pang+Lee+Vaithyanathan:02a} and \cite{Pang+Lee:04a} conducted by Pang and Lee. They utilized machine learning methods applied on the subjective parts of texts that were previously extracted, finding minimum cuts in graphs. They also released sentiment polarity dataset which contains 2000 movie reviews divided as \emph{positive} or \emph{negative}. Their work created the road-map for a series of similar studies that apply various techniques to the problem. 
In \cite{DBLP:conf/paclic/TsutsumiSE07} for example, they use a voting method of multiple classifiers for higher accuracy. Furthermore, in \cite{Singh2011}, authors go one step further, embedding a movie review classifier inside the hybrid movie recommender system they propose.
\par
Distributed word representations and deep learning models appeared recently in works like \cite{socher-EtAl:2013:EMNLP} where RNTNs (Recursive Neural Tensor Network) are presented, together with Stanford Sentiment Treebank, a dataset of \emph{positive} and \emph{negative} sentences extracted from movie reviews. RNTNs take as input a phrase of any length (sentences) and represent it through word vectors and a parse tree. The authors report that their new representation together with the phrase dataset boosted up the classification performance by 5.4\%. Other studies that followed (\cite{DBLP:journals/corr/KalchbrennerGB14}) presented implementations of even more advanced neural networks like Dynamic CNNs (Convolutional Neural Network).
A similar work that has relevance for us is \cite{maas-EtAl:2011:ACL-HLT2011} where authors present a log-bilinear vector space model capable of capturing sentiment and semantic similarities between words. Their model is based on a probabilistic distribution of documents that learns word representations in an unsupervised way. They used the dataset of Pang and Lee to evaluate their model and construct a larger movie review dataset of 50,000 movie reviews from IMBD. That dataset has been used in similar studies such as \cite{DBLP:conf/naacl/Johnson015} or \cite{pouransari2014deep} and is very popular in sentiment analysis. For our movie review experiments here, we also used this dataset so that we could directly compare our results with similar studies. For the phone review experiments, we used a dataset of unlocked mobile phones sold on Amazon. Users usually provide text comments, a 1~-~5 star rating or both for the products they buy online. This datasets of phones contains records with both star rating and text comments. We removed all reviews with a 3-star rating for a better separation of 1-star and 2-star reviews which were considered as \emph{negative}, from 4-star and 5-star reviews that were considered as \emph{positive}. A final dataset of 232546 reviews was reached. 
\section{Linguistic Regularity Benchmarks}
\label{linguisticRegularity}
One way for evaluating the quality of trained embedding models is to observe the semantic and syntactic regularities they manifest. This can be achieved by exercising the models in word analogy tasks of the form \dq{A is to B as C is to ?}. To give a correct answer the model needs to identify the missing word. The question is answered by finding the word D with representation $V_d$ closest to $V_b - V_a + V_c$ according to cosine similarity. 
We thus utilized the dataset of 19,544 questions also used by authors of \cite{pennington2014glove} for testing performance of Glove. It contains word analogy task questions created by Mikolov \emph{et. al.} in \cite{linguistic-regularities-in-continuous-space-word-representations}.
Semantic questions consist of gender (especially household) analogies like \dq{Brother is to Sister as Man is to ?}. In this case, the model must provide \dq{Woman} as the correct response. There is also a high number of questions like \dq{Rome is to Italy as Vienna is to ?} or \dq{Denmark is to Krone as Russia is to ?} for testing relations between countries, capitals, cities, currencies etc.  
Syntactic regularities are also important, especially in applications that work with part-of-speech relationships. They are evaluated with questions like \dq{Apparent is to Apparently as Quick is to ?} with \dq{Quickly} being the correct answer. Similar questions test comparative and superlative forms of adjectives, the adjective form of nationalities (e.g., \dq{England is to English as Italy is to ?}), present participle or past tense of verbs, plural forms of nouns etc. 
\par
We used these sets of questions to observe the role of corpus size, corpus theme and training method in syntactic and semantic regularities. For the same reason, we created and exercised 5-part cuts of our text bundles as well as fixed size (500 million tokens) versions of each of them. The basic attributes of all these text bundles are presented in Table~\ref{table:corporaCuts}. 
\begin{table}[ht] 
	\small 
	\centering       
	\setlength\tabcolsep{2.4pt}  
	\caption{Corpora of different sizes}  
	\begin{tabular}
		{l c c c | l c c c}  
		\toprule
		\textbf{Corpus Name}  & \textbf{Size} & \textbf{Voc} & \textbf{Dim} \enspace\enspace\quad & \quad \textbf{Corpus Name}  & \textbf{Size} & \textbf{Voc} & \textbf{Dim}  	\\ [0.1ex] 
		\midrule  
		Misch\_12B &  12B & 441415  & 300 \enspace\quad & \quad Review\_2.5B & 2.5B & 82682 & 300 		\\ [0.2ex]
		Misch\_24B & 24B & 1234020  & 300 \enspace\quad & \quad News\_250M & 250M & 79034 & 300					 \\ [0.2ex]
		Misch\_36B & 36B & 1730038 & 300 \enspace\quad & \quad News\_500M & 500M & 111606 & 300					 \\ [0.2ex]
		Misch\_48B & 48B & 1739498  & 300 \enspace\quad & \quad News\_750M & 750M & 141772 & 300					 \\ [0.2ex]
		Misch\_60B & 60B & 1748730 & 300 \enspace\quad & \quad News\_1B & 1B & 157522 & 300						 \\ [0.2ex]
		Lyrics\_5B & 5B & 41901 & 300 \enspace\quad & \quad News\_1.25B & 1.25B & 171143 & 300						 \\ [0.2ex]
		Lyrics\_10B & 10B & 45714 & 300 \enspace\quad & \quad Tweets\_100M & 100M & 49911 & 300						 \\ [0.2ex]
		Lyrics\_15B & 15B & 49294 & 300 \enspace\quad & \quad Tweets\_200M & 200M & 67154 & 300						 \\ [0.2ex]
		Lyrics\_20B & 20B & 53213 & 300 \enspace\quad & \quad Tweets\_300M & 300 & 78954 & 300					 \\ [0.2ex]
		Lyrics\_24B & 24B & 71590 & 300 \enspace\quad & \quad Tweets\_400M & 400M & 87529 & 300					 \\ [0.2ex]
		Review\_500M & 500M & 59745 & 300 \enspace\quad & \quad Tweets\_500M & 500M & 96055 & 300						 \\ [0.2ex]
		Review\_1B & 1B & 67547 & 300 \enspace\quad & \quad Misch\_500M & 500M & 112579 & 300					 \\ [0.2ex]
		Review\_1.5B & 1.5B & 71623 & 300 \enspace\quad & \quad Lyrics\_500M & 500M & 29305 & 300					 \\  [0.2ex]
		Review\_2B & 2B & 75084 & 300 \enspace\quad & \quad Text8Corpus & 17M & 71290 & 300			 \\			
		\bottomrule
	\end{tabular} 
	\label{table:corporaCuts}
\end{table}
For the word analogy task evaluations, we took 52 measures in total. The corresponding results are shown in Table~\ref{table:wordAnalogy}. In the first three columns, we provide attributes of each corpus like name, size of raw text and vocabulary size. In the fourth column \dq{Cov}, we present the coverage rate of each text set. It is the percentage of syntactic and semantic questions that were seen in that text. In columns 5 - 7 we show the semantic, syntactic and overall accuracy of Word2vec models of each corpus. The last 3 columns present the analogous accuracy metrics of Glove models. 
The first thing we notice from Table~\ref{table:wordAnalogy} is the low coverage rate that each corpus achieved. It is about 49\% in most of the cases. The only exceptions are Lyrics bundles (the worst) with coverage 44.91 or 47.92\% and the small \emph{Text8Corpus} (the best) that achieves an exceptionally high coverage rate of 91.21\%. As syntactic questions are highly covered by most of the corpora, overall accuracy that is the weighted average of syntactic and semantic accuracies, is always very close to the former. In fact, semantic questions are poorly covered. Questions about currencies or state-city relations were never seen in any text set. We see that coverage is influenced by vocabulary and theme of the texts. Song lyrics which are the most constrained thematically have small vocabulary (71590 the biggest) and thus the lowest coverage. On the other hand, \emph{Text8Corpus} of Wikipedia that is very small in size has a relatively rich vocabulary. Furthermore, as Wikipedia texts contain most of the terms comprised in the syntactic and semantic questions, it has the highest coverage rate. 
\par
The results indicate that semantic accuracy is higher than syntactic accuracy in almost every experiment. We also see a somehow proportional relation between size and vocabulary with the achieved score, at least on Tweets, News, Reviews and Misch texts. Actually, these text sets are similar to each other, containing texts of user discussions (with the exception of News). Their accuracy ranges from 36.11 to 61.32\% and grows with size and vocabulary. The five domain-specific Lyrics sets that contain texts of English songs have the lowest coverage and accuracy. Their vocabulary is small even though they are quite big in size. On the other hand, \emph{Text8Corpus} that is very small, reaches an accuracy of 29.65\% that is comparable to that of Tweets. 
The highest overall score is 61.32\% and was reached by \emph{Misch\_60B} that is our biggest corpus. This is lower than that of \emph{GoogleNews} which is reported to reach above 70 \% accuracy and 
cover almost 100\% of the questions (\url{https://code.google.com/archive/p/word2vec/}).
Regarding training method, it seems hard to find any pattern. We see that Word2vec performs better than Glove in Tweets, News and Text8Corpus which are small in size. It does better even in Lyrics that have the smallest vocabulary. Glove, on the other hand, performs better than Word2vec in reviews which have a medium size and also in Misch corpora that are the biggest, both in size and vocabulary.  
In summary, answering \hyperlink{rq1}{RQ1} we can say that based on our results, the best performance 
in semantic and syntactic word analogy tasks is achieved using multi-thematic texts (e.g., from Wikipedia), that are large in size and rich in vocabulary.                  
\begin{table}[ht] 
	\caption{Word analogy task accuracy scores}  
	\small 
	\centering      
	\setlength\tabcolsep{2.8pt}  
	\begin{tabular}
		{l | c c c | c c c | c c c}  
		\toprule
		& \multicolumn{9}{c}{\textbf{Properties} \qquad ~~~~ \qquad \textbf{Word2vec} \qquad ~~~~ \qquad \textbf{Glove }} \\
		\textbf{Corpus Name} & \textbf{Size} & \textbf{Voc} & \textbf{Cov} & \textbf{Sem} & \textbf{Sync} & \textbf{Overall} & \textbf{Sem} & \textbf{Sync} & \textbf{Overall}
		\\ [0.1ex] 
		\midrule
		Tweets\_100M & 100M & 44911 & 48.52 & 59.52  & 42.07 & 42.92 & 
		77.27  & 34.00 & 36.11 						 \\ [0.07ex] 
		Tweets\_200M &  200M & 67154 & 49.03 & 59.52  & 42.07 & 42.92 & 
		69.76  & 34.84 & 36.68 						 \\ [0.07ex] 
		Tweets\_300M & 300M & 78954 & 49.03 & 60.21 & 43.24 & 44.17 & 
		68.97 & 35.82 & 37.57						 \\ [0.07ex] 
		Tweets\_400M & 400M & 87529 & 49.03 & 60.77 & 44.12 & 44.87 & 
		68.18 & 37.30 & 38.93 						 \\ [0.07ex] 
		Tweets\_500M & 500M & 96055 & 49.03 & 60.77 & 44.32 & 45.05 & 
		69.17 & 38.22 & 39.86 						 \\ [0.07ex] 
		\hline
		News\_250M & 250M & 79034 & 49.03 & 54.74 & 45.30 & 45.79 & 
		37.15 & 33.57 & 33.67 						 \\ [0.07ex] 
		News\_500M & 500M & 111606 & 49.03 & 63.64 & 50.19 & 50.90 & 
		43.48 & 40.55 & 40.70 						 \\ [0.07ex] 
		News\_750M & 750M & 141772 & 49.03 & 73.12 & 51.32 & 52.47 & 
		62.06 & 44.50 & 45.43 						 \\ [0.07ex] 
		News\_1B & 1B & 157522 & 49.03 & 76.48 & 52.56 & 53.82 & 
		59.29 & 45.68 & 46.40 						 \\ [0.07ex] 
		News\_1.25B & 1.25B & 171143 & 49.03 & 76.88 & 52.39 & 53.68 & 
		66.40 & 45.86 & 46.94 						 \\ [0.07ex] 
		\hline
		Lyrics\_5B & 5B & 41901 & 44.91 & 21.9 & 8.78 & 9.41 & 
		7.14 & 1.76 & 2.02						 \\ [0.07ex] 
		Lyrics\_10B &  10B & 45714 & 44.91 & 25.71 & 9.72 & 10.48 & 
		12.14 & 1.35 & 1.87						 \\ [0.07ex] 
		Lyrics\_15B & 15B & 49294 & 44.91 & 22.62 & 9.34 & 9.98 & 
		9.76 & 1.30 & 1.71						 \\ [0.07ex] 
		Lyrics\_20B & 20B & 53213 & 44.92 & 24.41 & 9.61 & 10.32 & 
		9.00 & 1.62 & 1.97						 \\ [0.07ex] 
		Lyrics\_24B & 24B & 71590 & 47.92 & 26.29 & 9.77 & 10.59 & 
		10.34 & 2.02 & 2.43					 \\ [0.07ex] 
		\hline
		Review\_500M & 500M & 59745 & 48.74 & 85.38 & 48.97 & 50.90 & 
		84.19 & 49.11 & 50.98					 \\ [0.07ex] 
		Review\_1B & 1B & 67547 & 49.03 & 87.15 & 49.93 & 51.90 & 
		89.13 & 53.09 & 54.99					 \\ [0.07ex] 
		Review\_1.5B & 1.5B & 71623 & 49.03 & 88.14 & 52.04 & 53.04 & 
		90.32 & 55.73 & 57.56					 \\ [0.07ex] 
		Review\_2B & 2B & 75084 & 49.03 & \textbf{88.14} & 54.11 & 55.91 & 
		\textbf{90.32} & 57.64 & 59.36					 \\ [0.07ex] 
		Review\_2.5B & 2.5B & 82682 & 49.03 & 83.20 & 56.82 & 58.21 & 
		86.96 & 57.35 & 58.91					 \\ [0.07ex] 
		\hline
		Misch\_12B & 12B & 441415 & 49.03 & 79.23 & \textbf{59.22} & 59.48 & 
		81.42 & 59.77 & 60.92					 \\ [0.07ex] 
		Misch\_24B & 24B & 1234020 & 49.03 & 79.23 & 57.34 & 57.96 & 
		84.39 & 55.81 & 57.32					 \\ [0.07ex] 
		Misch\_36B & 36B & 1730038 & 49.04 & 80.26 & 58.66 & 57.97 & 
		79.72 & 56.99 & 58.19					 \\ [0.07ex] 
		Misch\_48B & 48B & 1739498 & 49.04 & 81.11 & 58.66 & 59.05 & 
		84.45 & 59.32 & 60.65					 \\ [0.07ex] 
		Misch\_60B & \textbf{60B} & \textbf{1748730} & 49.04 & 82.39 & 59.04 & \textbf{59.68} &  
		85.04 & \textbf{59.99} & \textbf{61.32}					 \\ [0.07ex] 
		Text8Corpus & 17M & 71290 & \textbf{91.21} & 39.25  & 40.61 & 40.05 & 
		40.40  & 22.00 & 29.65 						 \\
		\bottomrule
	\end{tabular} 
	\label{table:wordAnalogy}
\end{table}
\section{Role of Corpus Characteristics}
\label{corpusCharacteristics}
In this section, we experiment with word embeddings obtained from training corpora of different sizes and themes, trained using Glove and Word2vec methods. Our intention is to observe any influence that those factors might have on the quality of embeddings and thus classification performance on each of the tasks of Section~\ref{tasksAndDatasets}. To observe the significance of those factors we also perform statistical hypothesis testing on the results. For this reason, we set the following four null hypotheses: 
\begin{description}
	\itemsep0.7em  
	\item[$\mathbf{H_0}$] \emph{There is no significant difference between accuracy scores of word embedding features derived from different text collections.}
	\item[$\mathbf{Hs_0}$] \emph{There is no significant difference between accuracy scores of word embedding features derived from text corpora of different sizes.}
	\item[$\mathbf{Hm_0}$] \emph{There is no significant difference between accuracy scores of word embedding features trained with Word2vec and those trained with Glove.}
	\item[$\mathbf{Ht_0}$] \emph{There is no significant difference between accuracy score results of word embedding features derived from text corpora of different topics.}
\end{description}
As for the significance level, we chose $\alpha = 0.05$ which is typical of this type of studies. We performed paired t-test analysis on five samples of each score, comparing the different accuracy values in each experimental set of the following subsections. 
\subsection{Scores of all Models}
\label{bigCorporaScores}
\begin{figure}[ht]
	\centering
	\begin{minipage}{.5\columnwidth}
		\centering
		\includegraphics[width=0.97\columnwidth]{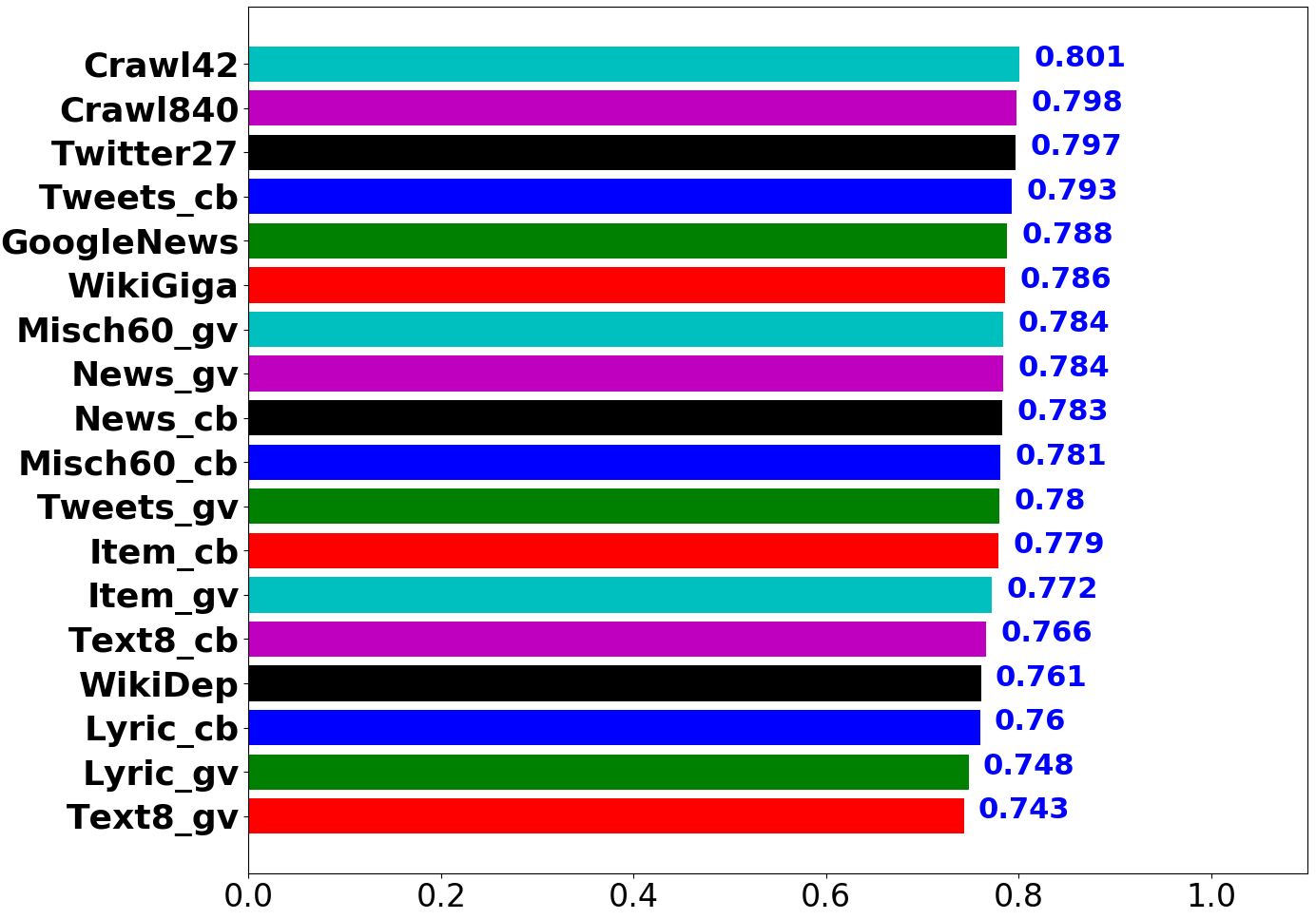}
		\caption{All models on Twitter tweets}
		\label{fig:allOnTweets}
	\end{minipage}%
	\begin{minipage}{.5\columnwidth}
		\centering
		\includegraphics[width=0.97\columnwidth]{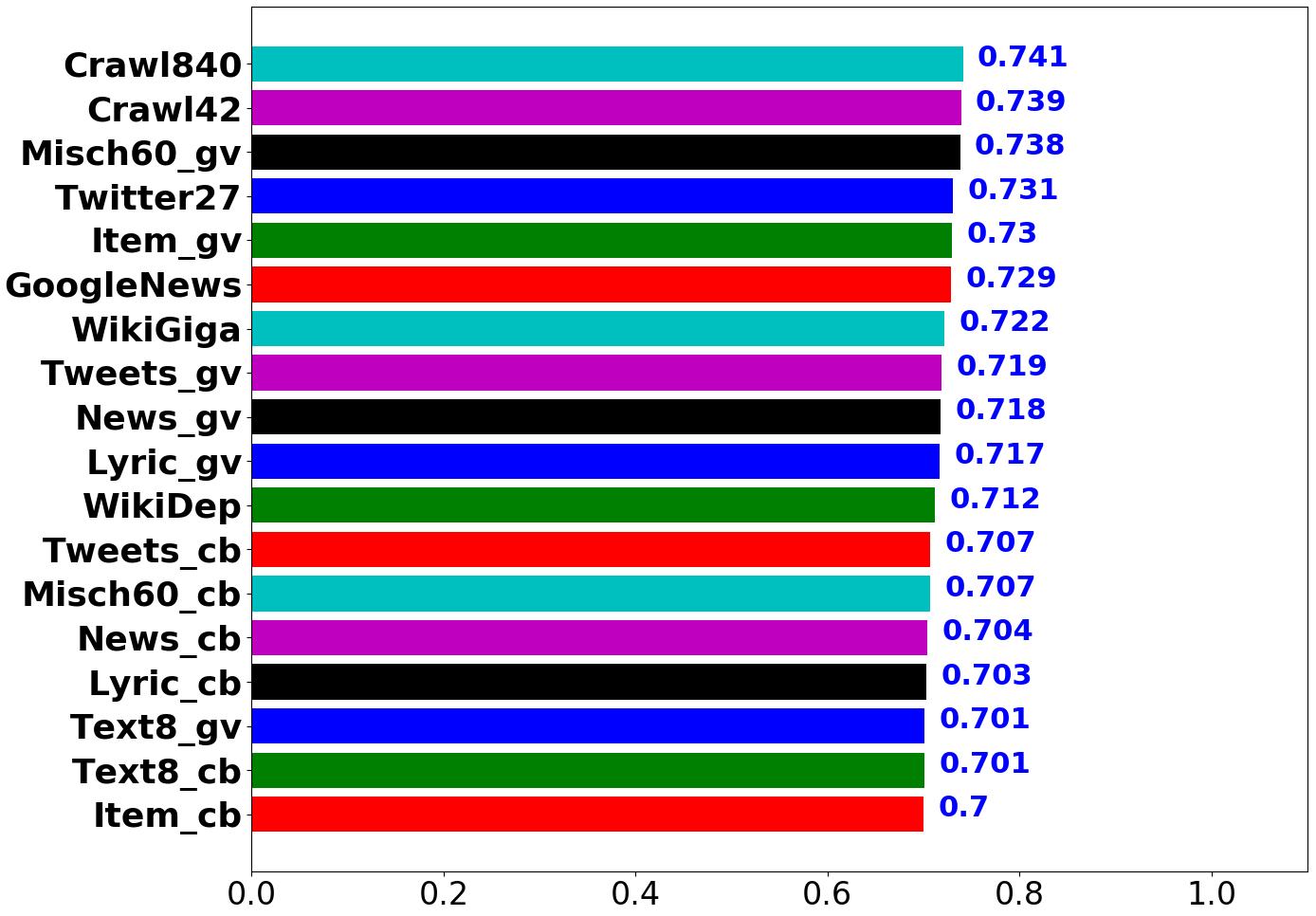}
		\caption{All models on song lyrics}
		\label{fig:allOnLyrics}
	\end{minipage}
\end{figure}

\begin{figure}[ht]
	\centering
	\begin{minipage}{.5\columnwidth}
		\centering
		\includegraphics[width=0.97\columnwidth]{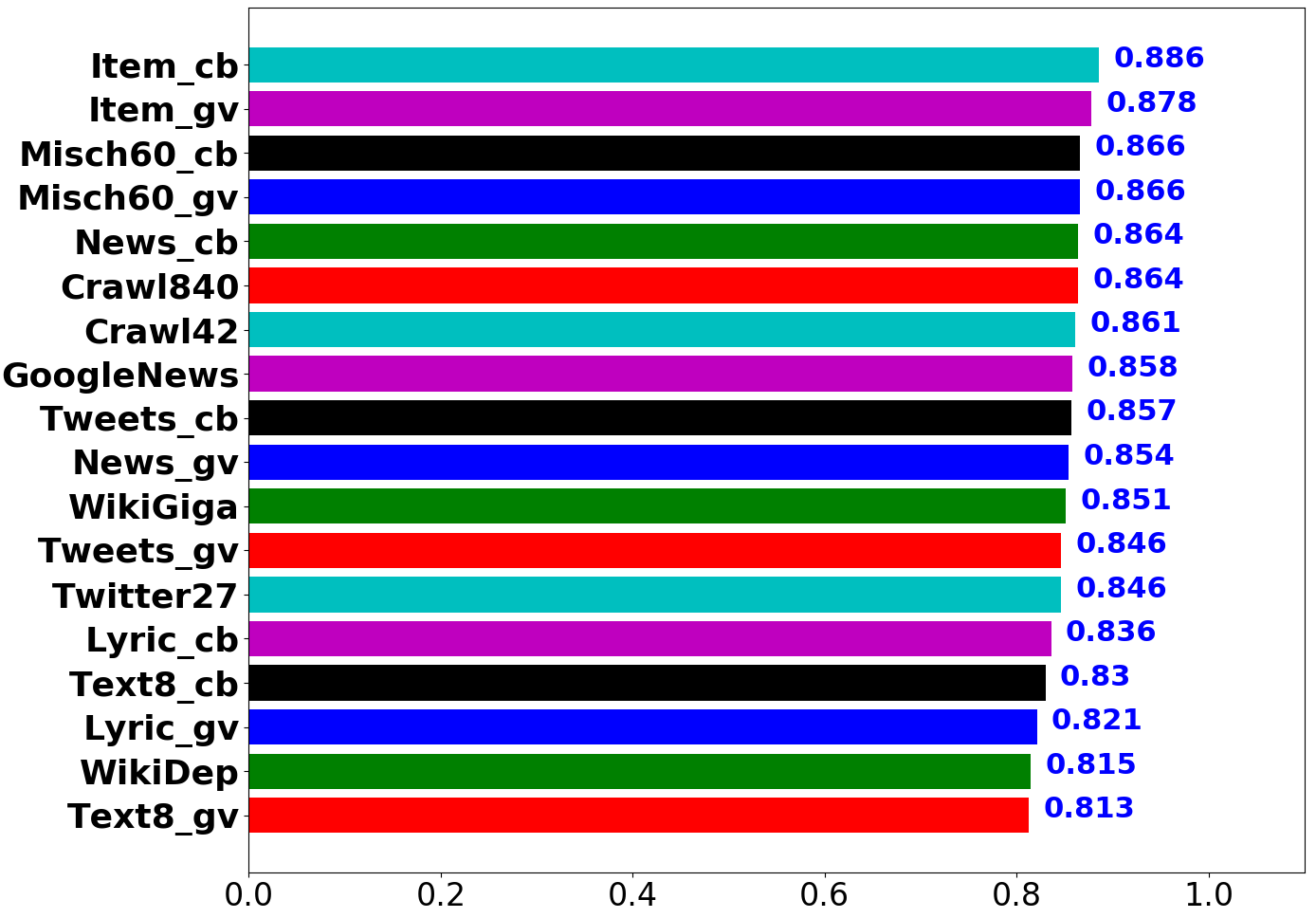}
		\caption{All models on movie reviews}
		\label{fig:allOnMovies}
	\end{minipage}%
	\begin{minipage}{.5\columnwidth}
		\centering
		\includegraphics[width=0.97\columnwidth]{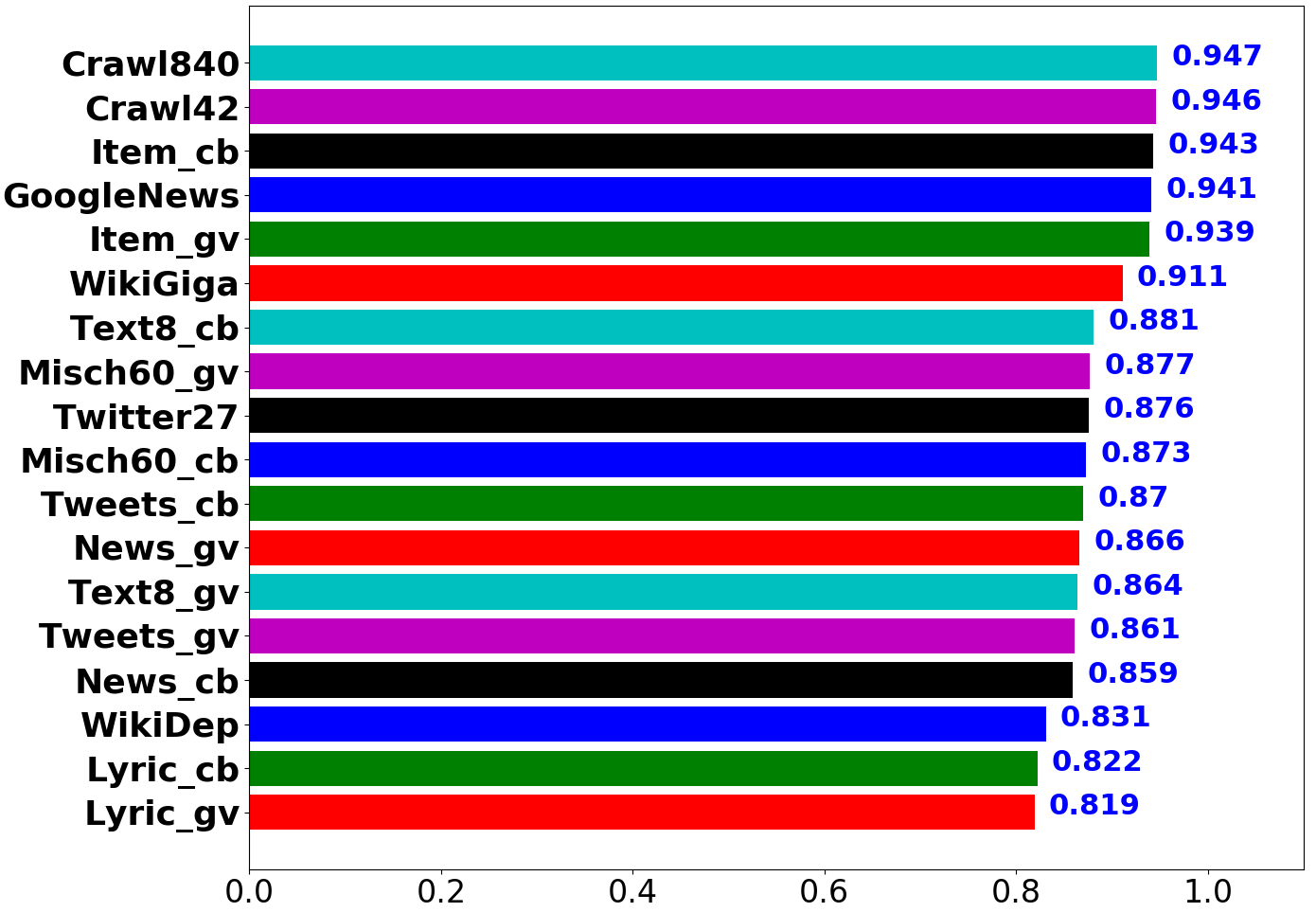}
		\caption{All models on phone reviews}
		\label{fig:allOnPhones}
	\end{minipage}
\end{figure}

In this first set of experiments we tried all the 18 models of Table~\ref{table:corporaProperties} on the four sentiment analysis tasks. The point here is to spot the best models, see what results they provide, and compare those results with other similar studies using same datasets. Statistical tests will also show us how confident we are for rejecting $H_0$ and confirm that word embeddings trained on different corpora do yield diverse accuracy scores on the chosen tasks. Figures~\ref{fig:allOnTweets} - \ref{fig:allOnPhones} summarize obtained 5-fold cross-validation accuracies. 
We see that on the classification of tweets, the top score is that of \emph{Crawl42} which reaches slightly more than 80\%. It is closely followed by \emph{Crawl840} (biggest corpus) and \emph{Twitter27} (biggest tweet bundle). Moreover, our collections of tweets (\emph{Tweets\_gv} and \emph{Tweets\_cb}) perform well (only 1.7\% lower than top score) even though they are considerably smaller in size. The worst are \emph{Lyric} models and \emph{Text8\_gv} (the smallest) with a margin of  0.058 from the top. Macro F1 of the top model is 0.793 which is about 7\% lower than that of other state-of-the-art works (see Section~\ref{tweetSent}). Nevertheless, we believe that this scoring gap is more related to the dataset and the classifier we utilized rather than to the embeddings. 
\par
On the classification of song lyrics, we got a similar ranking. Once again \emph{Crawl840} and \emph{Crawl42} top the list with top accuracy that is about 74\%. Our \emph{Misch60} bundle follows closely together with \emph{Twitter27}. \emph{Item} models perform well also, despite their relatively small size. We are surprised to see that the two lyrics models \emph{Lyric\_gv} and \emph{Lyric\_cb} are well behind, positioned in the second half of the list, even though they contain song lyrics and are big in size. Another surprising result is the considerable gap between Glove and Word2vec versions of the models we trained ourself. This indicates that Glove might be considerably better than Word2vec on lyrics. The top results we got on song lyrics is somehow low. It is yet comparable with analogous literature reports such as \cite{zhong2012music}. 
\par
Accuracy scores on movie reviews are presented in Figure~\ref{fig:allOnMovies}. Here we see that our \emph{Item} collection of reviews performs best with top score reaching about 89\%. It is followed by our \emph{Misch} and \emph{News} bundles. The biggest corpora are positioned in the middle and the small \emph{Text8} ends the list scoring less than 82\%. Due to the popularity of IMDB movie review dataset, there are a lot of recent studies with which to compare. Actually the result we got here is similar with that of several other studies. In \cite{maas-EtAl:2011:ACL-HLT2011} for example, authors of IMDB dataset achieved 88\% accuracy (almost what we got here) utilizing a probabilistic model they developed. Also in \cite{DBLP:conf/naacl/Johnson015}, they feed bag-of-3-grams features in a CNN of three layers, reaching an accuracy of 92\% which is even higher than ours. 
\par
Accuracy scores of all models on phone reviews are presented in Figure~\ref{fig:allOnPhones}. We see that \emph{Crawl840} and \emph{Crawl42} bundles are again on top with an accuracy score of almost 95\%. Also, our \emph{Item} models are doing well, same as on movie reviews. They are positioned $3^{rd}$ and $5^{th}$. The two lyrics bundles are the worst, with scores lower than 83\%. This may be due to their small vocabulary size. Something else to notice here is the considerable difference of about 0.13 between the top and the bottom of the table. That difference is the highest among the four experiments. We did not find any literature results on the phone dataset. 
We performed paired t-test analysis, comparing the values of the two best models against the two worst in each of the four experiments. The resulting $t$ and $p$ values are listed in Table~\ref{table:rejectingH0}. As we can see, in every experiment, $t$ values are much greater than $p$ values. Furthermore, $p$ values are much smaller than the chosen value of significance ($\alpha = 0.05$). As a result, we have evidence to reject hypothesis $H_0$ with very high confidence. 
\begin{table}[ht]
	\caption{Comparing highest and lowest scores of each experiment}
		\centering
	\label{table:rejectingH0}
	\begin{tabular}{l | c c c l}
		\toprule
		~~Compared Models & Task & t & p & ~~~Verdict  \\ [0.1ex]
		\midrule
		Crawl42 vs Text8\_gv & Tweets & 5.26 & 0.00077 & $H_0$ rejected   \\ [0.22ex]
		Crawl840 vs Lyric\_gv & Tweets & 5.04 & 0.00092 & $H_0$ rejected   \\ [0.22ex]
		Crawl840 vs Item\_cb & Lyrics & 4.11 & 0.00341 & $H_0$ rejected		\\ [0.22ex]
		Crawl42 vs Text8\_cb & Lyrics & 3.921 & 0.0042 & $H_0$ rejected		\\ [0.22ex]
		Item\_cb vs Text8\_gv & Movies & 5.36 & 0.00068 & $H_0$ rejected   \\ [0.22ex]
		Item\_gv vs WikiDep & Movies & 5.11 & 0.00081 & $H_0$ rejected   	\\ [0.22ex]
		Crawl840 vs Lyric\_gv & Phones & 8.37 & 0.000011 & $H_0$ rejected   \\ [0.22ex]
		Crawl42 vs Lyric\_cb & Phones & 7.56 & 0.000025 & $H_0$ rejected   	\\ [0.22ex]
		\bottomrule
	\end{tabular}
\end{table}
The different word embedding collections do perform differently from each other. Top scores we got here are slightly weaker than state-of-art models on each of the four sentiment analysis tasks. Also, the biggest models in terms of size and vocabulary richness usually provide top results. Among them, \emph{Crawl840} and \emph{Crawl42} are obviously the best. Another interesting result is the good performance of item review bundles (\emph{Item\_cb} and \emph{Item\_gv}) on sentiment analysis of movie and phone reviews. This might be an indication of a high thematic relevance of training text corpora on those two tasks. In Secton~\ref{scoresFixedSize} we specifically address this issue.  
\subsection{Scores of Misch Corpora Size Cuts}
\label{scoresOnMischCuts}
\begin{figure}[ht]
	\centering
	\begin{minipage}{.5\columnwidth}
		\centering
		\includegraphics[width=0.97\columnwidth]{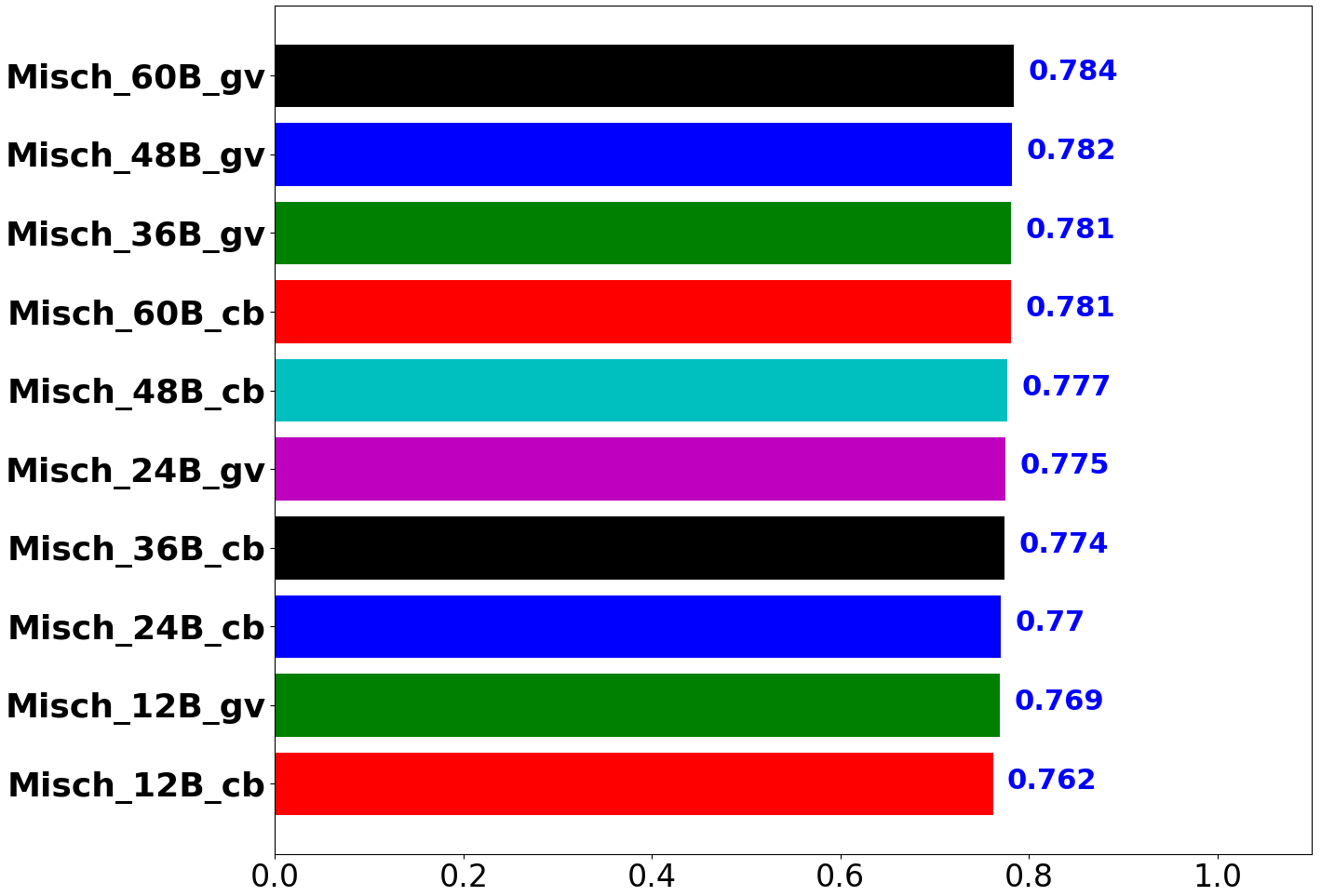}
		\caption{Misch models on Twitter tweets}
		\label{fig:mischOnTweets}
	\end{minipage}%
	\begin{minipage}{.5\columnwidth}
		\centering
		\includegraphics[width=0.97\columnwidth]{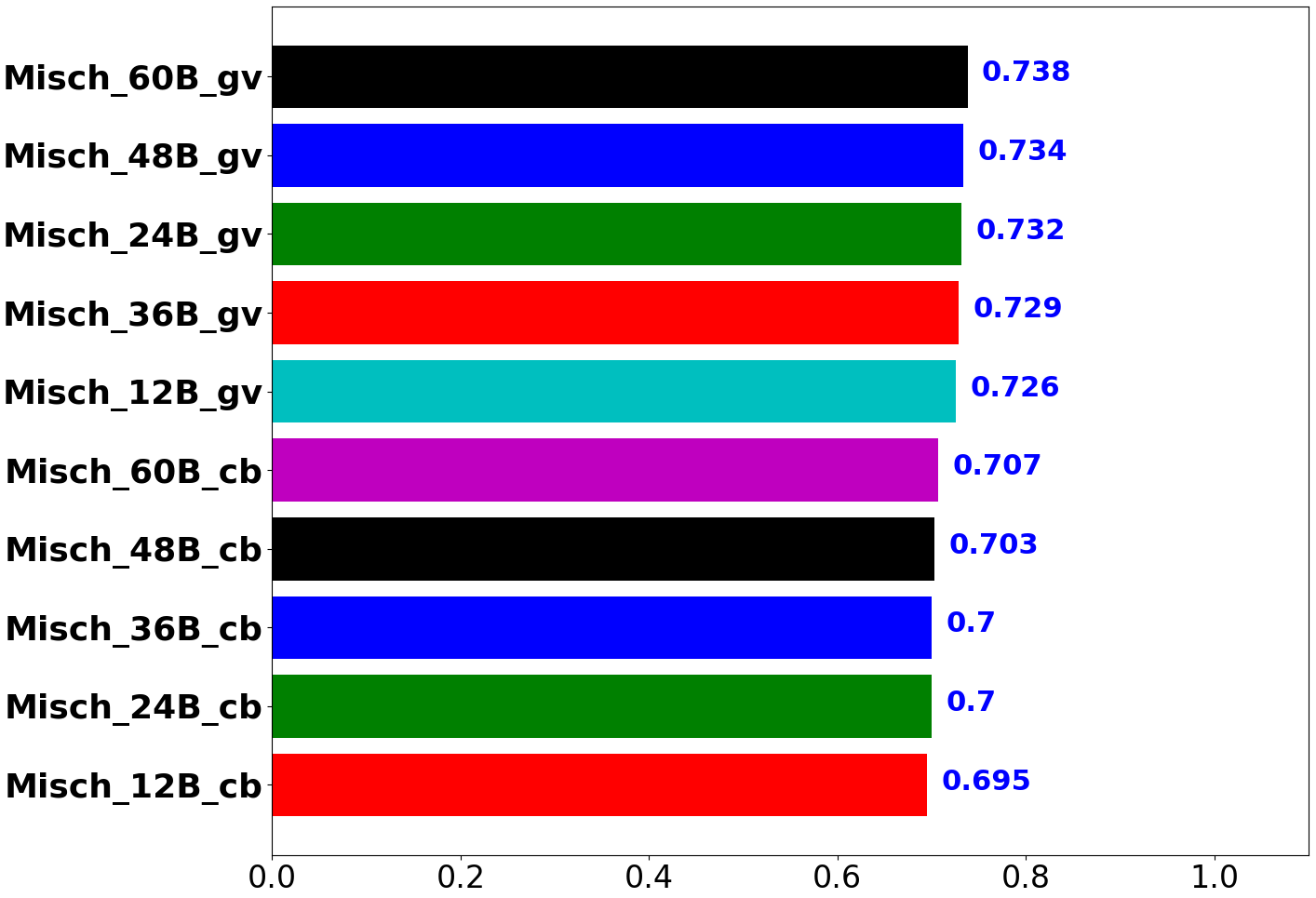}
		\caption{Misch models on song lyrics}
		\label{fig:mischOnLyrics}
	\end{minipage}
\end{figure}

\begin{figure}[ht]
	\centering
	\begin{minipage}{.5\columnwidth}
		\centering
		\includegraphics[width=0.97\columnwidth]{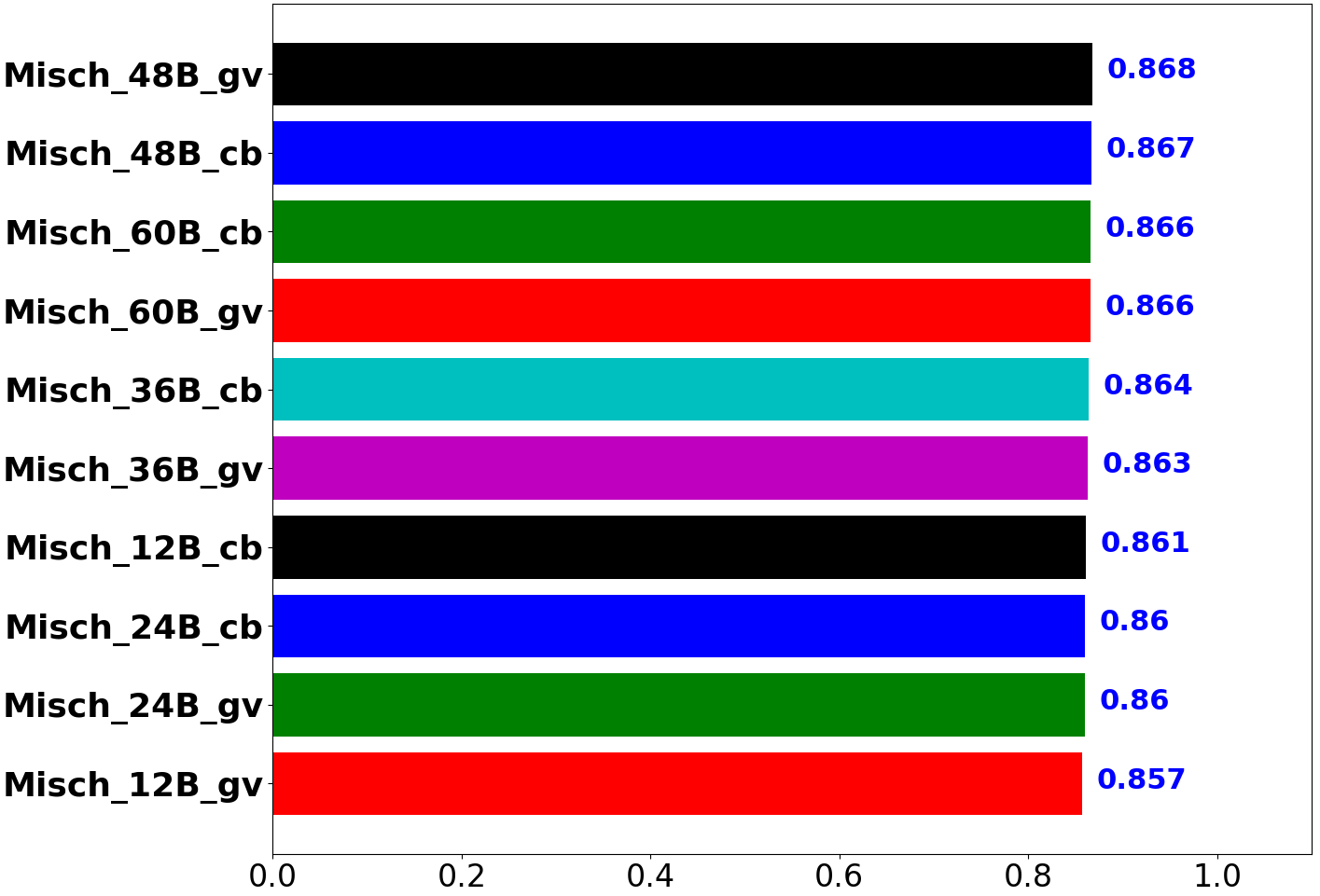}
		\caption{Misch models on movie reviews}
		\label{fig:mischOnMovies}
	\end{minipage}%
	\begin{minipage}{.5\columnwidth}
		\centering
		\includegraphics[width=0.97\columnwidth]{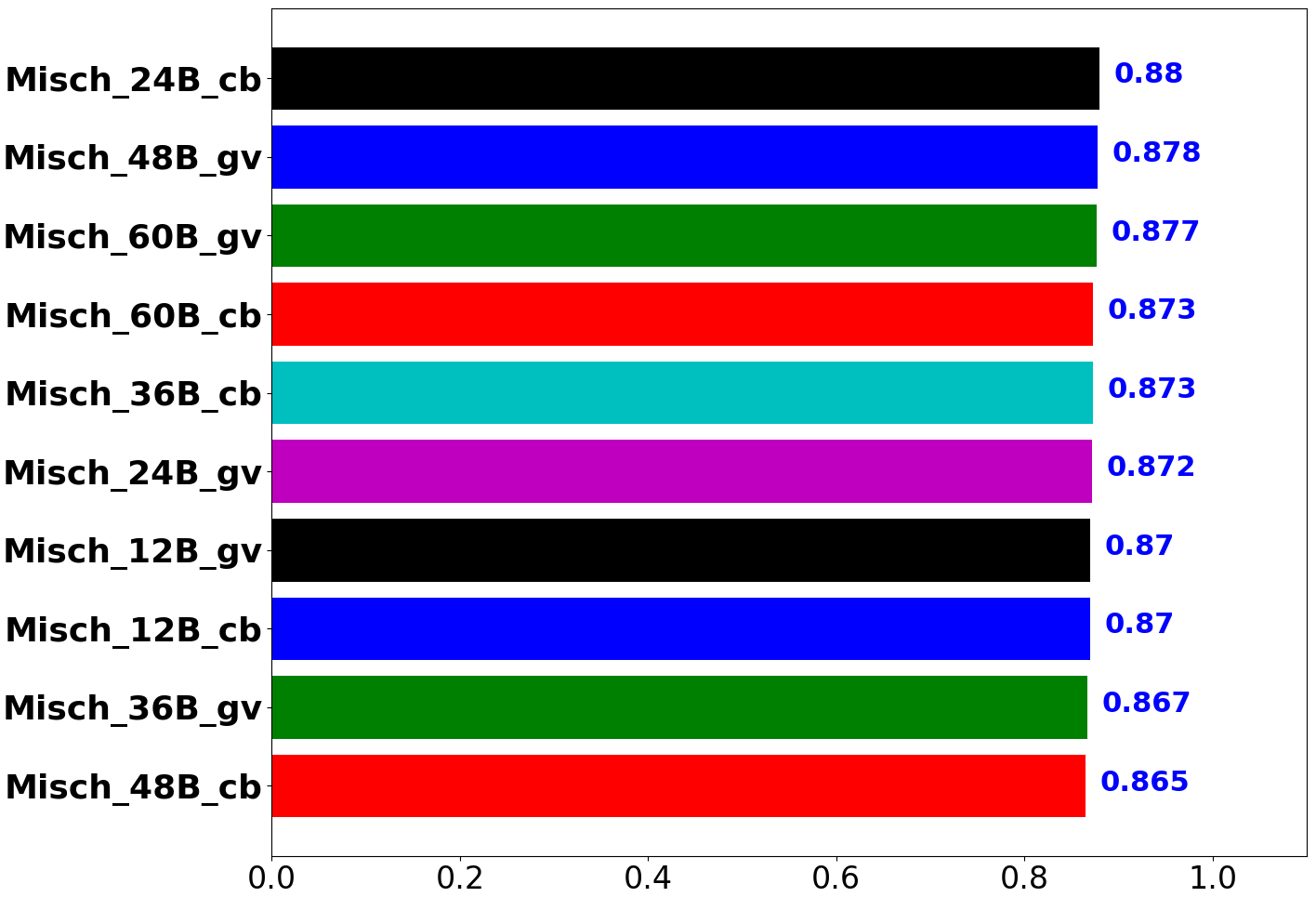}
		\caption{Misch models on phone reviews}
		\label{fig:mischOnPhones}
	\end{minipage}
\end{figure}

Here we try to observe any role that training corpus size and training method might have on the performance of word embeddings. This way we can assess $Hs_0$ and $Hm_0$ hypothesis listed at the beginning of this section. To evade any influence of text topic, we exercise the five cuts of the same text bundle (MischBundle) trained with Glove and Word2vec (ten models in total). Figures~\ref{fig:mischOnTweets} - \ref{fig:mischOnPhones} report 5-fold cross-validation accuracies of each model. 
On tweet datasets, we got the results that are shown in Figure~\ref{fig:mischOnTweets}. The best model is \emph{Misch\_60B\_gv} and the worst one is \emph{Misch\_12B\_cb}. We see that accuracy scores are close with each other, with a difference of only 0.022 from first to last. There is still a visible advantage of Glove over Word2vec, for every couple of scores of the same size. Furthermore, it seems that bigger cuts always outperform the smaller ones of the same method. At the end of this section, we provide statistical evidence to support these observations. 
In Figure~\ref{fig:mischOnLyrics} we see a similar picture on song lyrics. Once again Glove models top the list, followed by Word2vec models. Furthermore, models scores are ranked in accordance with their training corpus size. The score difference between first and last models is 0.043 which is bigger than the one in tweets. We can also spot a considerable difference in accuracy between the worst Glove model and the best Word2vec one. This is an indication of a strong effect of training method on this task. 
The results on movie and phone reviews (Figures~\ref{fig:mischOnMovies} and \ref{fig:mischOnMovies}) are similar with each other but very different from those in tweets and lyrics. Accuracy scores are very close here, with extreme differences of only 0.011 and 0.015 respectively. Also, we do not observe any regular tendency regarding the effect of training method or corpus size. 
To further validate these preliminary insights, we performed statistical analysis on score samples of every experimental set. For the experiments on tweets and lyrics, we compared accuracy samples of Glove and Word2vec models of the same size. This way we could assess $Hm_0$. As for $Hs_0$, scores of biggest and smallest models of each method were contrasted. Obtained statistical results are presented in Tables~\ref{table:Hs0andHm0OnTweets} and \ref{table:Hs0andHm0OnLyrics}. In the first five rows of these two tables, we see that $t$ values are much greater than $p$ values. Also $p$ is smaller than $\alpha$ (0.05) in every case. As a result, we can confidently reject $Hm_0$ and confirm that indeed, training method of word embeddings has a significant effect on their quality on sentiment analysis of tweets and song lyrics. We also see that this effect is bigger on song lyrics and that Glove is better than Word2vec in every case. Regarding $Hs_0$, $t$ and $p$ values in the last two rows of both tables indicate that it can also be rejected. We also see that corpus size effect is stronger on tweets and somehow weaker on song lyrics. 
Accuracy scores on movie and phone reviews are very close to each other. Regarding $Hm_0$, we compared the models of different sizes trained with Glove and Word2vec. We also compared biggest and smallest models of each training method. Obtained results are listed in Tables~\ref{table:Hs0andHm0OnMovies} and \ref{table:Hs0andHm0OnPhones}. As we can see, both $Hm_0$ and $Hs_0$ could not be rejected in almost every comparison. We can thus infer that training method and training corpus size of word embeddings have no significant effect on their quality on sentiment analysis of movie and phone reviews. 
\begin{table}[ht]
	\caption{Assessing $Hs_0$ and $Hm_0$ with Misch models on tweets}
		\centering
	\begin{tabular}{l | c c l}
		\toprule
		~~~~~~~~~~~Compared Models & t & p & ~~~~Verdict  \\ [0.1ex]
		\midrule
		Misch\_60B\_gv vs Misch\_60B\_cb & 2.62 & 0.031 & $Hm_0$ rejected   \\ [0.3ex]
		Misch\_48B\_gv vs Misch\_48B\_cb & 2.81 & 0.022 & $Hm_0$ rejected	\\ [0.3ex]	
		Misch\_36B\_gv vs Misch\_36B\_cb & 2.55 & 0.034 & $Hm_0$ rejected   \\ [0.3ex]
		Misch\_24B\_gv vs Misch\_24B\_cb & 3.29 & 0.011 & $Hm_0$ rejected   \\ [0.3ex]
		Misch\_12B\_gv vs Misch\_12B\_cb & 3.14 & 0.013 & $Hm_0$ rejected   \\ [0.3ex]
		Misch\_60B\_gv vs Misch\_12B\_gv & 4.64 & 0.0011 & $Hs_0$ rejected   \\ [0.3ex]
		Misch\_60B\_cb vs Misch\_12B\_cb & 4.85 & 0.0012 & $Hs_0$ rejected   \\ [0.3ex]
		\bottomrule
	\end{tabular}
	\label{table:Hs0andHm0OnTweets}
\end{table}
\begin{table}[ht]
	\caption{Assessing $Hs_0$ and $Hm_0$ with Misch models on lyrics}
		\centering
	\begin{tabular}{l | c c l}
		\toprule
		~~~~~~~~~~~Compared Models & t & p & ~~~~Verdict  \\ [0.1ex]
		\midrule
		Misch\_60B\_gv vs Misch\_60B\_cb & 5.22 & 0.00081 & $Hm_0$ rejected   \\ [0.3ex]
		Misch\_48B\_gv vs Misch\_48B\_cb & 5.12 & 0.00091 & $Hm_0$ rejected	\\ [0.3ex]	
		Misch\_36B\_gv vs Misch\_36B\_cb & 5.07& 0.00097 & $Hm_0$ rejected   \\ [0.3ex]
		Misch\_24B\_gv vs Misch\_24B\_cb & 4.89 & 0.0012 & $Hm_0$ rejected   \\ [0.3ex]
		Misch\_12B\_gv vs Misch\_12B\_cb & 4.96 & 0.0011 & $Hm_0$ rejected   \\ [0.3ex]
		Misch\_60B\_gv vs Misch\_12B\_gv & 3.69 & 0.0061 & $Hs_0$ rejected   \\ [0.3ex]
		Misch\_60B\_cb vs Misch\_12B\_cb & 2.46 & 0.0394 & $Hs_0$ rejected   \\ [0.3ex]
		\bottomrule
	\end{tabular}
	\label{table:Hs0andHm0OnLyrics}
\end{table}
\begin{table}[]
	\caption{Assessing $Hs_0$ and $Hm_0$ with Misch models on movies}
		\centering
	\begin{tabular}{l | c c l}
		\toprule
		~~~~~~~~~~~Compared Models & t & p & ~~~~Verdict  \\ [0.1ex]
		\midrule
		Misch\_60B\_gv vs Misch\_60B\_cb & -0.86 & 0.414 & $Hm_0$ not rejected   \\ [0.3ex]
		Misch\_48B\_gv vs Misch\_48B\_cb & 0.96 & 0.365 & $Hm_0$ not rejected	\\ [0.3ex]	
		Misch\_36B\_gv vs Misch\_36B\_cb & -1.12& 0.295 & $Hm_0$ not rejected   \\ [0.3ex]
		Misch\_24B\_gv vs Misch\_24B\_cb & -0.66 & 0.527 & $Hm_0$ not rejected   \\ [0.3ex]
		Misch\_12B\_gv vs Misch\_12B\_cb & -1.49 & 0.174 & $Hm_0$ not rejected   \\ [0.3ex]
		Misch\_60B\_gv vs Misch\_12B\_gv & 2.27 & 0.053 & $Hs_0$ not rejected   \\ [0.3ex]
		Misch\_60B\_cb vs Misch\_12B\_cb & 2.09 & 0.07 & $Hs_0$ not rejected   \\ [0.3ex]
		\bottomrule
	\end{tabular}
	\label{table:Hs0andHm0OnMovies}
\end{table}
\begin{table}[]
	\caption{Assessing $Hs_0$ and $Hm_0$ with Misch models on phones}
		\centering
	\begin{tabular}{l | c c l}
		\toprule
		~~~~~~~~~~~Compared Models & t & p & ~~~~Verdict  \\ [0.1ex]
		\midrule
		Misch\_60B\_gv vs Misch\_60B\_cb & 1.04 & 0.328 & $Hm_0$ not rejected   \\ [0.3ex]
		Misch\_48B\_gv vs Misch\_48B\_cb & 2.38 & 0.044 & $Hm_0$ rejected	\\ [0.3ex]	
		Misch\_36B\_gv vs Misch\_36B\_cb & -1.45 & 0.185 & $Hm_0$ not rejected   \\ [0.3ex]
		Misch\_24B\_gv vs Misch\_24B\_cb & -1.72 & 0.123 & $Hm_0$ not rejected   \\ [0.3ex]
		Misch\_12B\_gv vs Misch\_12B\_cb & 0.94 & 0.3741 & $Hm_0$ not rejected   \\ [0.3ex]
		Misch\_60B\_gv vs Misch\_12B\_gv & 2.06 & 0.260 & $Hs_0$ not rejected   \\ [0.3ex]
		Misch\_60B\_cb vs Misch\_12B\_cb & 2.15 & 0.063 & $Hs_0$ not rejected   \\ [0.3ex]
		\bottomrule
	\end{tabular}
	\label{table:Hs0andHm0OnPhones}
\end{table}
\subsection{Scores of Domain-specific Size Cuts}
\label{scoresOnDomain}
\begin{figure}[ht]
	\centering
	\begin{minipage}{.5\columnwidth}
		\centering
		\includegraphics[width=0.97\columnwidth]{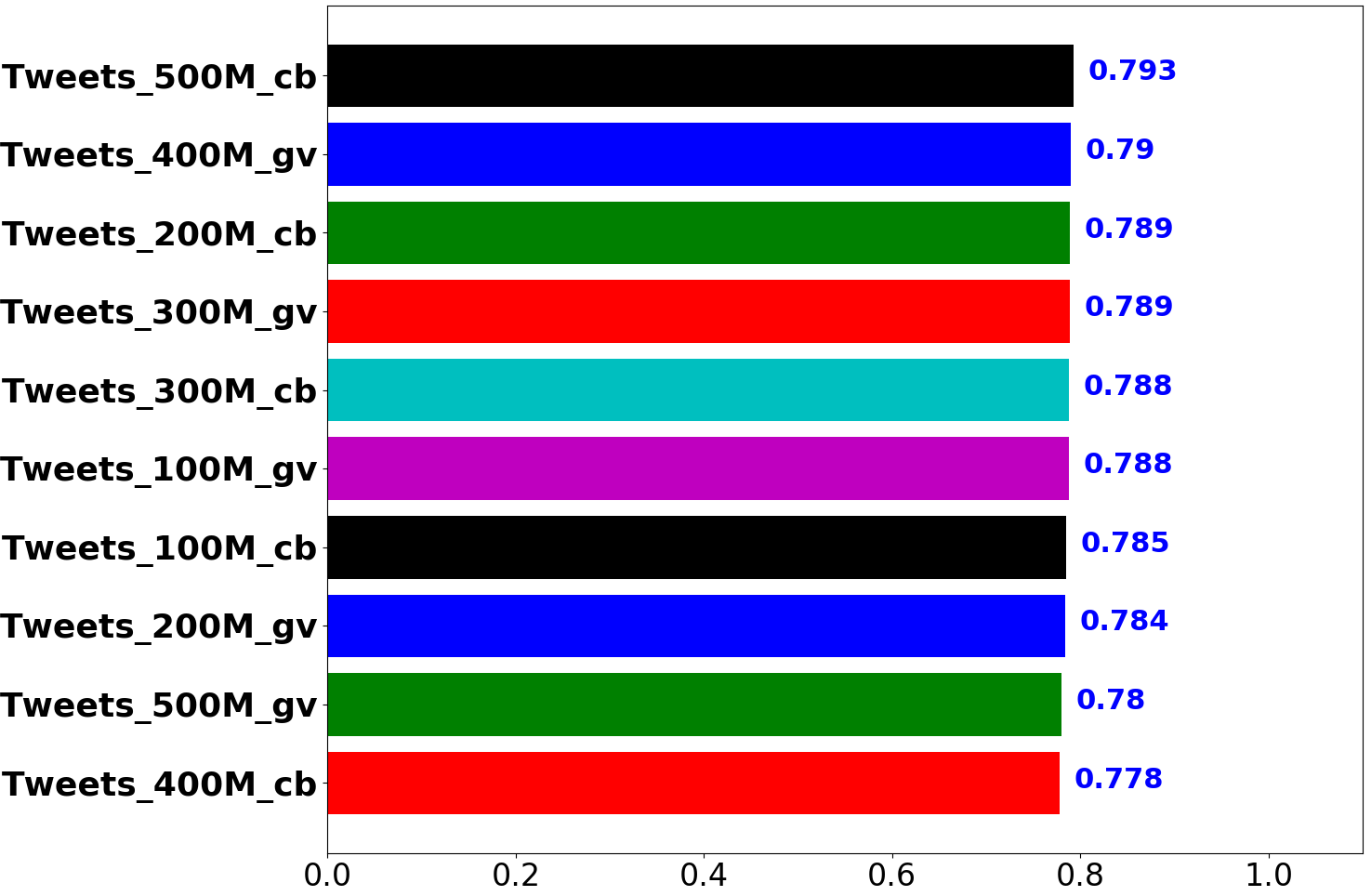}
		\caption{Tweet models on Twitter tweets}
		\label{fig:tweetsOnTweets}
	\end{minipage}%
	\begin{minipage}{.5\columnwidth}
		\centering
		\includegraphics[width=0.97\columnwidth]{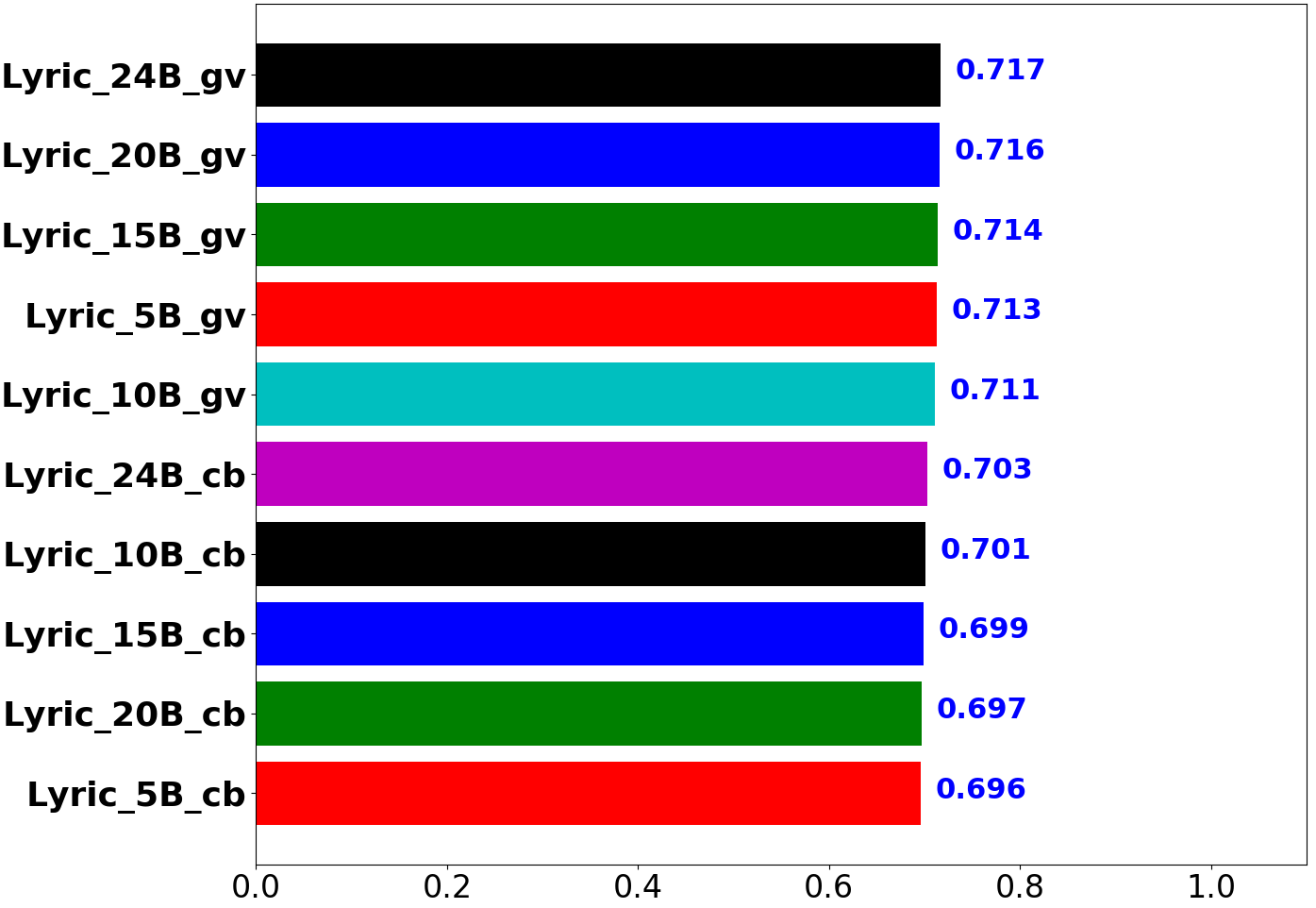}
		\caption{Lyric models on song lyrics}
		\label{fig:lyricsOnLyrics}
	\end{minipage}
\end{figure}

\begin{figure}[!t]
	\centering
	\begin{minipage}{.5\columnwidth}
		\centering
		\includegraphics[width=0.97\columnwidth]{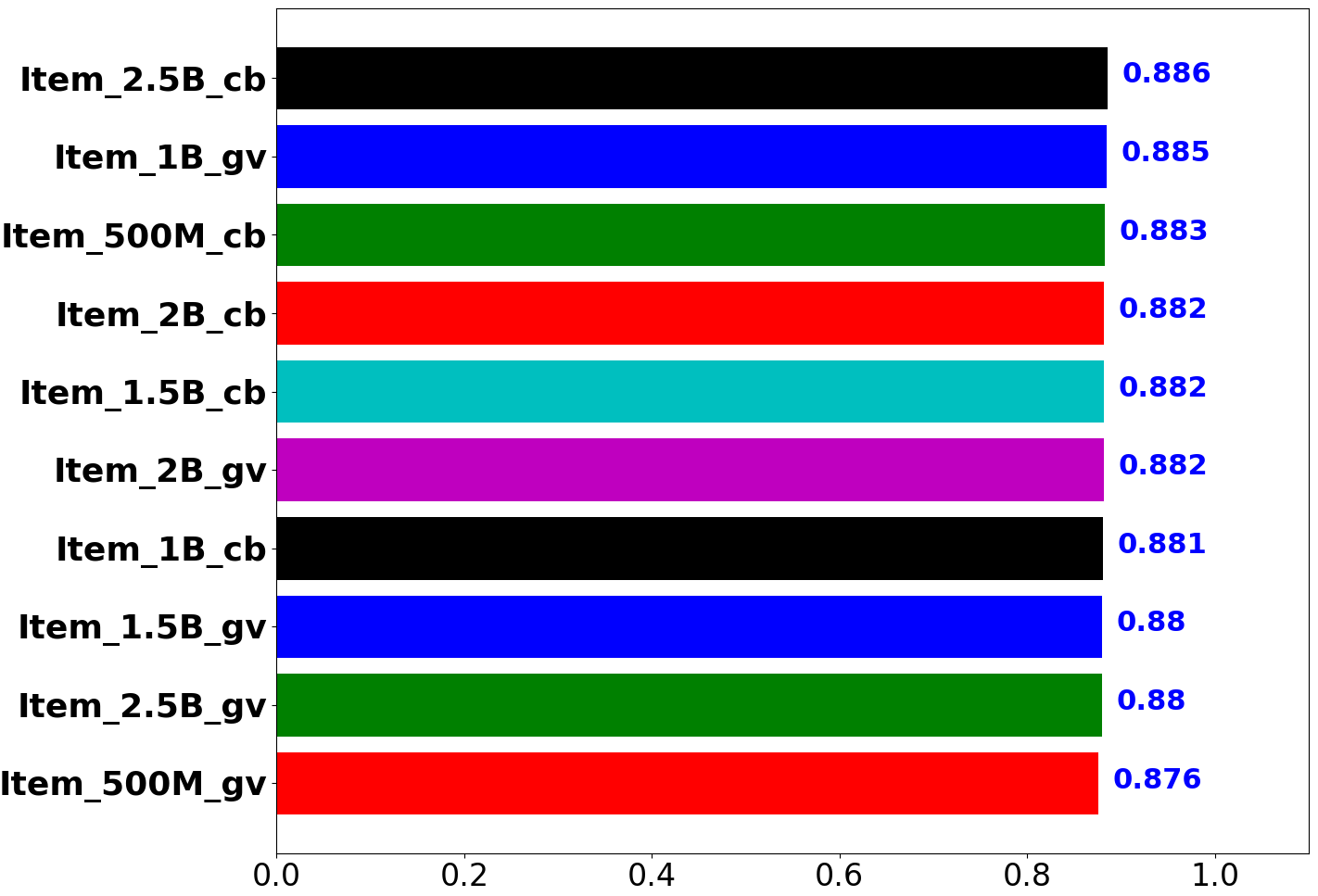}
		\caption{Review models on movie reviews}
		\label{fig:revsOnMovies}
	\end{minipage}%
	\begin{minipage}{.5\columnwidth}
		\centering
		\includegraphics[width=0.97\columnwidth]{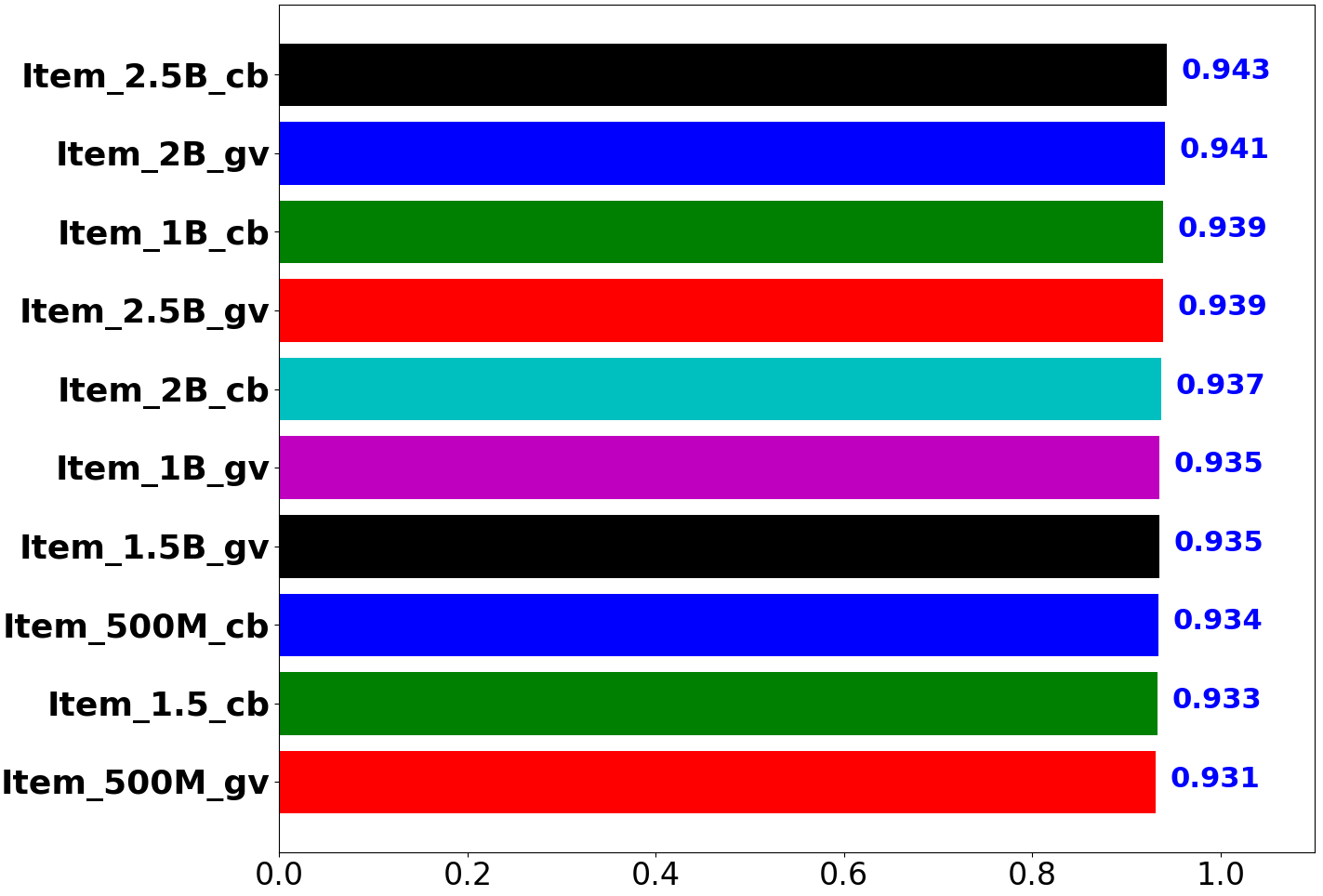}
		\caption{Review models on phone reviews}
		\label{fig:revsOnItems}
	\end{minipage}
\end{figure}

Same as in the previous subsection, in this set of experiments we also try to observe the role of model size on the performance of word embeddings. This time we utilize the five size cuts of the four domain-specific corpora on each dataset of that same domain. Accuracy scores are presented in Figures~\ref{fig:tweetsOnTweets} - \ref{fig:revsOnItems}.  
Accuracy scores on tweets dataset are higher than those of Misch models in the previous subsection. This might be an indication of corpus thematic influence. The values are however strongly constrained in an interval of 0.015. The best model is \emph{Tweets\_500M\_cb} and the worst is \emph{Tweets\_400M\_cb}. Regarding size and training method, it seems hard to spot any tendency.  
On song lyrics, we got scores that are lower than those of Misch models. Lyric\_24B\_gv which is the best models reaches a score of 0.717 only. However, results follow the same tendency. We see bigger models outperform the smaller counterparts. Also, Glove models provide better results than Word2vec ones. In fact, there is a visible gap of at least 0.008 between them. 
On movie reviews (Figure~\ref{fig:revsOnMovies}), we see that \emph{Item} models perform much better than \emph{Misch} models in the previous section. There is a difference of 0.018 between the two top models of the corresponding experiments. This is a strong indication of corpus thematics influence on sentiment analysis of movie reviews. The scores of the different cuts and methods are however very similar to each other, with a difference of only 0.01 from top to bottom.    
On phone reviews, we see the same tendency as in movie reviews. Top item review models (again \emph{Item\_2.5B\_cb}) yields an accuracy of 0.943 which is 0.063 higher than the score of  \emph{Misch\_24B\_cb}. Same as in movie reviews, this result suggests a strong influence of corpus thematics on sentiment analysis of phone reviews. Also, the models of different sizes and training methods are again very close to each other. 
Same as in the previous section, we compared Glove models with Word2vec corresponding models of the same size using paired t-test analysis. We also compared biggest and smallest models of both methods. Considering obtained $p$ values and $\alpha = 0.05$ that we set at the beginning of this section, we found that $Hm_0$ and $Hs_0$ could not be rejected in most of the cases (Table~\ref{table:Hs0andHm0OnTweetsFromTweets}). It means that effects of size and training method are not significant in most of the experiments with tweets. This is something strange considering the results with Misch models in the previous section. A possible explanation for this could be the fact that size differences of training corpora here are much smaller (few million tokens) compared with those of \emph{Misch} models (many billion tokens). 
\begin{table}[!t]
	\caption{Assessing $Hs_0$ and $Hm_0$ with Tweet models on tweets}
		\centering
	\begin{tabular}{l | c c l}
		\toprule
		~~~~~~~~~~~Compared Models & t & p & ~~~~Verdict  \\ [0.1ex]
		\midrule
		Tweets\_500M\_gv vs Tweets\_500M\_cb & 1.94 & 0.088 & $Hm_0$ not rejected   \\ [0.3ex]
		Tweets\_400M\_gv vs Tweets\_400M\_cb & 1.97 & 0.084 & $Hm_0$ not rejected	\\ [0.3ex]	
		Tweets\_300M\_gv vs Tweets\_300M\_cb & 2.33 & 0.048 & $Hm_0$ rejected   \\ [0.3ex]
		Tweets\_200M\_gv vs Tweets\_200M\_cb & 2.16 & 0.062 & $Hm_0$ not rejected   \\ [0.3ex]
		Tweets\_100M\_gv vs Tweets\_100M\_cb & 2.51 & 0.036 & $Hm_0$ rejected   \\ [0.3ex]
		Tweets\_500M\_gv vs Tweets\_100M\_gv & 2.14 & 0.064 & $Hs_0$ not rejected   \\ [0.3ex]
		Tweets\_500M\_cb vs Tweets\_100M\_cb & 1.66 & 0.135 & $Hs_0$ not rejected   \\ [0.3ex]
		\bottomrule
	\end{tabular}
	\label{table:Hs0andHm0OnTweetsFromTweets}
\end{table}
\begin{table}[!t]
	\caption{Assessing $Hs_0$ and $Hm_0$ with Lyric models on lyrics}
		\centering
	\begin{tabular}{l | c c l}
		\toprule
		~~~~~~~~~~~Compared Models & t & p & ~~~~Verdict  \\ [0.1ex]
		\midrule
		Lyric\_24B\_gv vs Lyric\_24B\_cb & 2.21 & 0.058 & $Hm_0$ not rejected   \\ [0.3ex]
		Lyric\_20B\_gv vs Lyric\_20B\_cb & 2.41 & 0.042 & $Hm_0$ rejected	\\ [0.3ex]	
		Lyric\_15B\_gv vs Lyric\_15B\_cb & 2.43 & 0.041 & $Hm_0$ rejected   \\ [0.3ex]
		Lyric\_10B\_gv vs Lyric\_10B\_cb & 2.59 & 0.032 & $Hm_0$ rejected   \\ [0.3ex]
		Lyric\_5B\_gv vs Lyric\_5B\_cb & 2.38 & 0.044 & $Hm_0$ rejected   \\ [0.3ex]
		Lyric\_24B\_gv vs Lyric\_5B\_gv & 2.75 & 0.025 & $Hs_0$ rejected   \\ [0.3ex]
		Lyric\_24B\_cb vs Lyric\_5B\_cb & 2.83 & 0.022 & $Hs_0$ rejected   \\ [0.3ex]
		\bottomrule
	\end{tabular}
	\label{table:Hs0andHm0OnLyricsFromLyrics}
\end{table}
Statistical results of song lyrics accuracies are listed in Table~\ref{table:Hs0andHm0OnLyricsFromLyrics}. Same as in the previous section, we compared Glove models with Word2vec corresponding models of the same size. As we can see, $Hs_0$ is rejected in both comparisons whereas $Hm_0$ is rejected in four of the five. These results are in line with those of the previous section and confirm that training method and training corpus size do affect the quality of word embeddings when used on sentiment analysis of lyrics.   
\begin{table}[!t]
	\caption{Assessing $Hs_0$ and $Hm_0$ with Item models on movies}
		\centering
	\begin{tabular}{l | c c l}
		\toprule
		~~~~~~~~~~~Compared Models & t & p & ~~~~Verdict  \\ [0.1ex]
		\midrule
		Item\_2.5B\_gv vs Item\_2.5B\_cb & 2.21 & 0.058 & $Hm_0$ not rejected   \\ [0.3ex]
		Item\_2B\_gv vs Item\_2B\_cb & 2.16 & 0.062 & $Hm_0$ not rejected	\\ [0.3ex]	
		Item\_1.5B\_gv vs Item\_1.5B\_cb & 1.86 & 0.1 & $Hm_0$ not rejected   \\ [0.3ex]
		Item\_1B\_gv vs Item\_1B\_cb & 2.07 & 0.072 & $Hm_0$ not rejected   \\ [0.3ex]
		Item\_500M\_gv vs Item\_500M\_cb & 1.98 & 0.083 & $Hm_0$ not rejected   \\ [0.3ex]
		Item\_2.5B\_gv vs Item\_500M\_gv & 2.03 & 0.076 & $Hs_0$ not rejected   \\ [0.3ex]
		Item\_2.5B\_cb vs Item\_500M\_cb & 1.79 & 0.111 & $Hs_0$ not rejected   \\ [0.3ex]
		\bottomrule
	\end{tabular}
	\label{table:Hs0andHm0OnMoviesFromItems}
\end{table}
\begin{table}[!t]
	\caption{Assessing $Hs_0$ and $Hm_0$ with Item models on phones}
		\centering
	\begin{tabular}{l | c c l}
		\toprule
		~~~~~~~~~~~Compared Models & t & p & ~~~~Verdict  \\ [0.1ex]
		\midrule
		Item\_2.5B\_gv vs Item\_2.5B\_cb & 1.82 & 0.106 & $Hm_0$ not rejected   \\ [0.3ex]
		Item\_2B\_gv vs Item\_2B\_cb & 1.63 & 0.141 & $Hm_0$ not rejected	\\ [0.3ex]	
		Item\_1.5B\_gv vs Item\_1.5B\_cb & 1.99 & 0.081 & $Hm_0$ not rejected   \\ [0.3ex]
		Item\_1B\_gv vs Item\_1B\_cb & 2.04 & 0.075 & $Hm_0$ not rejected   \\ [0.3ex]
		Item\_500M\_gv vs Item\_500M\_cb & 2.15 & 0.063 & $Hm_0$ not rejected   \\ [0.3ex]
		Item\_2.5B\_gv vs Item\_500M\_gv & 2.06 & 0.073 & $Hs_0$ not rejected   \\ [0.3ex]
		Item\_2.5B\_cb vs Item\_500M\_cb & 2.2   & 0.059 & $Hs_0$ not rejected   \\ [0.3ex]
		\bottomrule
	\end{tabular}
	\label{table:Hs0andHm0OnPhonesFromItems}
\end{table}
Tables~\ref{table:Hs0andHm0OnMoviesFromItems} and \ref{table:Hs0andHm0OnPhonesFromItems} list statistical results of comparisons on movie and phone reviews. We see that $p$ values are always greater than 0.05 in all comparisons of both experimental sets. Thus, $Hs_0$ and $Hm_0$ cannot be rejected. We could not observe any significant influence of training method or corpus size on movie or phone reviews. 
\subsection{Scores of Fixed-size Domain-specific Corpora}
\label{scoresFixedSize}
\begin{figure}[ht]
	\centering
	\begin{minipage}{.5\columnwidth}
		\centering
		\includegraphics[width=0.97\columnwidth]{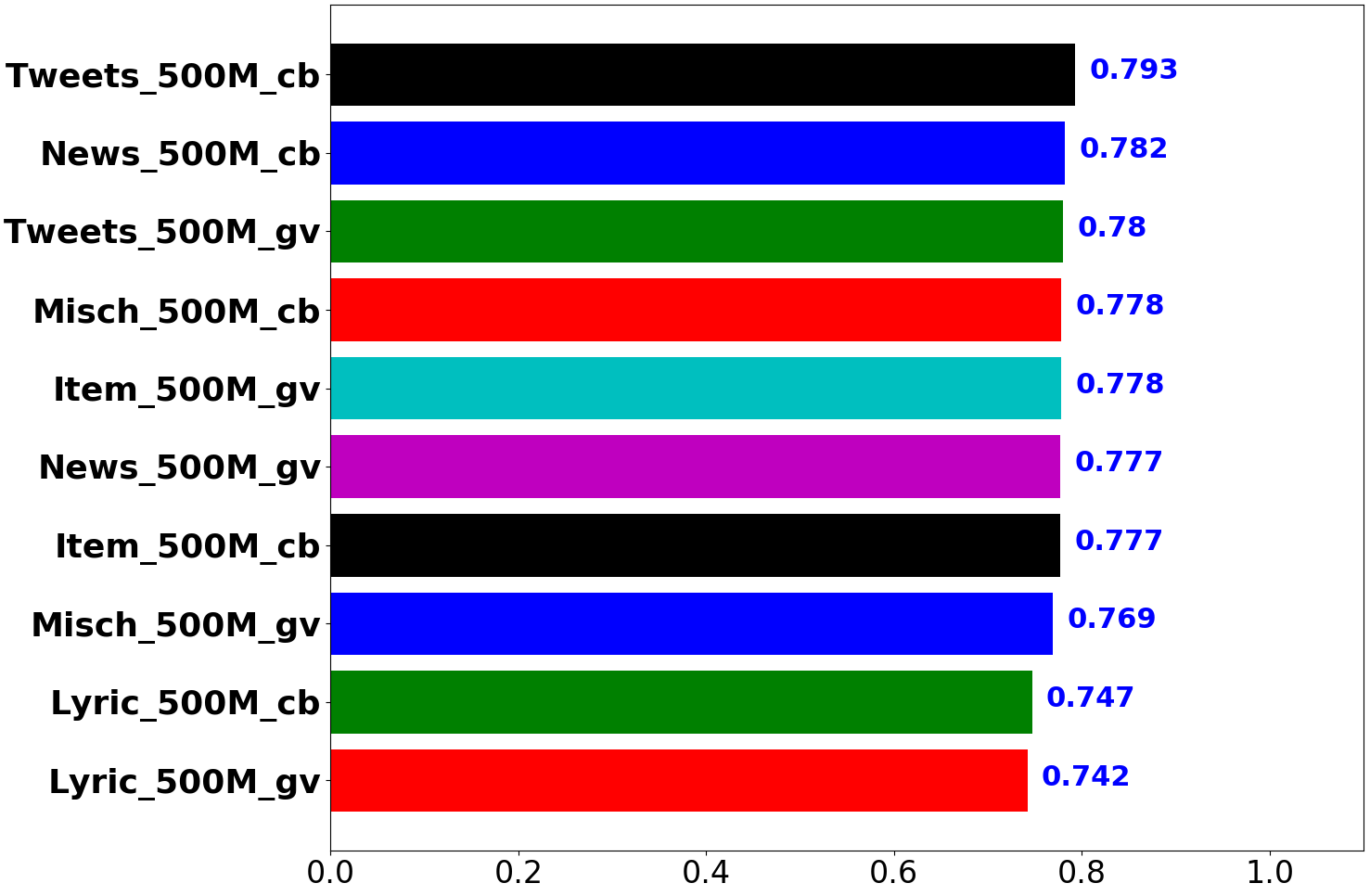}
		\caption{500M models on Twitter tweets}
		\label{fig:500mOnTweets}
	\end{minipage}%
	\begin{minipage}{.5\columnwidth}
		\centering
		\includegraphics[width=0.97\columnwidth]{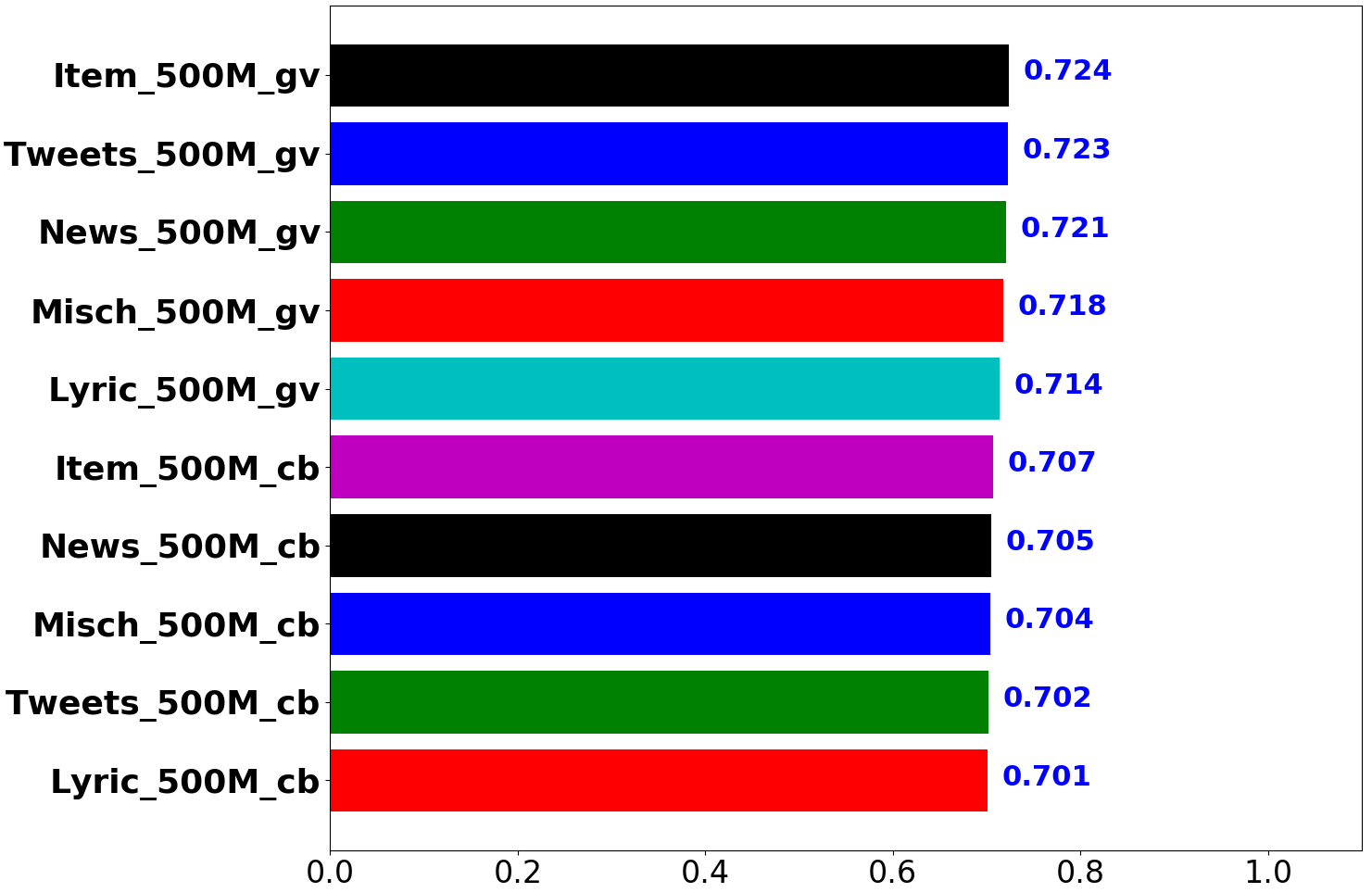}
		\caption{500M models on song lyrics}
		\label{fig:500mOnLyrics}
	\end{minipage}
\end{figure}

\begin{figure}[ht]
	\centering
	\begin{minipage}{.5\columnwidth}
		\centering
		\includegraphics[width=0.97\columnwidth]{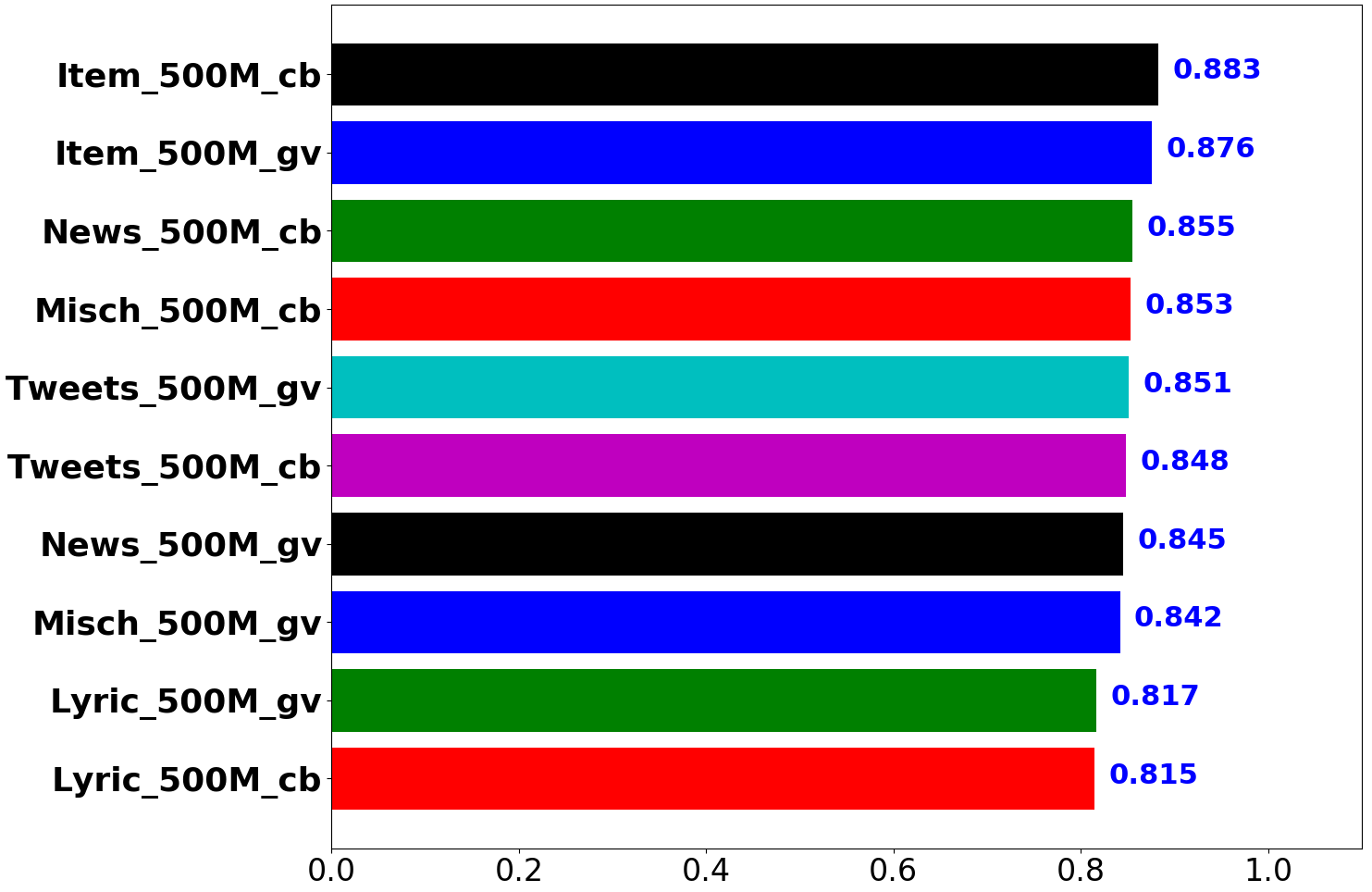}
		\caption{500M models on movie reviews}
		\label{fig:500mOnMovies}
	\end{minipage}%
	\begin{minipage}{.5\columnwidth}
		\centering
		\includegraphics[width=0.97\columnwidth]{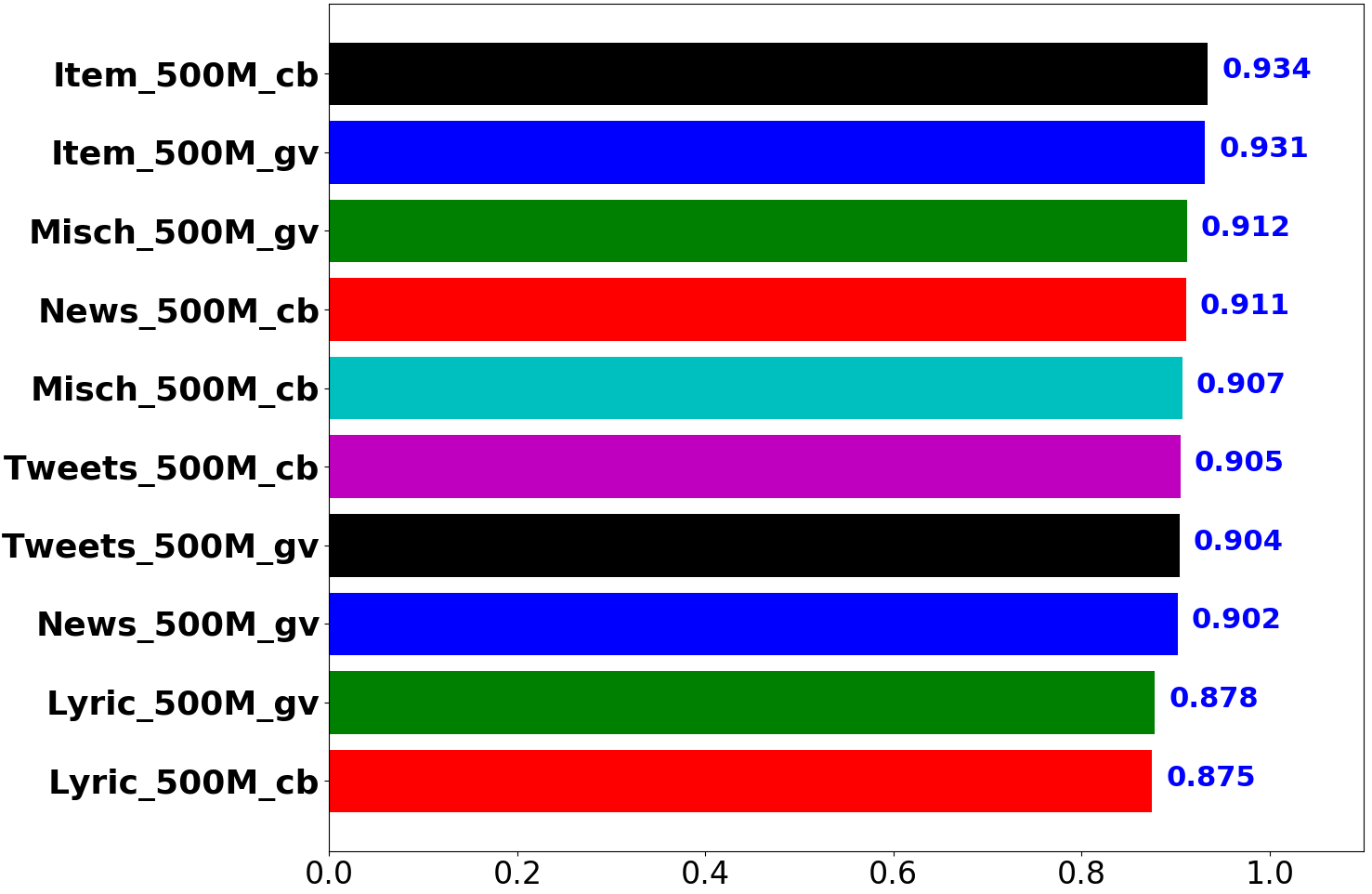}
		\caption{500M models on news articles}
		\label{fig:500mOnItems}
	\end{minipage}
\end{figure}

In this last set of experiments, we assess the performance of five equally-sized text corpora cuts trained with Glove and Word2vec on the four sentiment analysis tasks. The goal here is to observe any influence of corpus thematic relevance in quality of embeddings and evaluate hypothesis $Ht_0$. The results we got are shown in Figures~\ref{fig:500mOnTweets} - \ref{fig:500mOnItems}. 
In Figure~\ref{fig:500mOnTweets} we see that the two models of tweets are doing pretty well on Twitter sentiment analysis, positioned in the first and third place. Scores of other domain corpora like news and item reviews are positioned in the middle of the list. The two lyrics corpora are listed at the bottom with a considerable disadvantage of more than 0.022. Figure~\ref{fig:500mOnLyrics} summarizes results obtained on song lyrics. Here we see a controversial situation. From one point, the five Glove models are distinctively better Word2vec ones, same as in the previous corresponding experiments on lyrics. On the other hand, we see the two lyrics models performing worse than every other model trained with the same method. Obtained accuracy scores are also close in value, with a maximal difference of only 0.023 from the top to the bottom.    
\par
Results on movie reviews are presented in Figure~\ref{fig:500mOnMovies}. Here we see a distinct superiority of the two \emph{Item} models over the rest. There is also a distinct inferiority of the two \emph{Lyric} models, same as in the previous experiments. Results of Figure~\ref{fig:500mOnItems} on phone reviews are very similar. We have \emph{Item} model heading the list with a considerable advantage over the rest. \emph{Tweets}, \emph{News} and \emph{Misch} models are position in the middle of the list. As in the other experiments, \emph{Lyric} models perform badly.  
\begin{table}
	\caption{Assessing $Ht_0$ hypothesis}
		\centering
	\label{table:rejectingHt0}
	\begin{tabular}{l | c c c l}
		\toprule
		~~~~~~~~~~~~~Compared Models & Task & t & p & ~~~~Verdict  \\ [0.1ex]
		\midrule
		Tweets\_500M\_cb vs Misch\_500M\_cb & Tweets & 2.89 & 0.02 & $Ht_0$ rejected   \\ [0.3ex]
		Tweets\_500M\_gv vs Misch\_500M\_gv & Tweets & 3.24 & 0.011 & $Ht_0$ rejected\\ [0.3ex]
		Item\_500M\_cb vs Misch\_500M\_cb & Movies & 4.56 & 0.0018 & $Ht_0$ rejected   \\ [0.3ex]
		Item\_500M\_gv vs Misch\_500M\_gv & Movies & 4.27& 0.0027 & $Ht_0$ rejected   \\ [0.3ex]
		Item\_500M\_cb vs Misch\_500M\_cb & Phones & 4.12 & 0.0033 & $Ht_0$ rejected   \\ [0.3ex]
		Item\_500M\_gv vs Misch\_500M\_gv & Phones & 5.02 & 0.001 & $Ht_0$ rejected   \\ [0.3ex]
		\bottomrule
	\end{tabular}
\end{table}
A quick look at the results of the four experiments indicates that there is a thematic influence of corpora on accuracy scores obtained in tweets, phone and movie reviews. Results on song lyrics are very similar in value. Also, the two domain-specific models (\emph{Lyric\_500M\_cb} and \emph{Lyric\_500M\_gv}) perform worse than the other models. For these reasons, we performed t-test analysis on tweets, phones, and movies to see if we could reject or not $Ht_0$ hypothesis. We compared the two best performing models of each domain with the two models of \emph{Misch} corpus which is thematically neutral. The results are presented in Table~\ref{table:rejectingHt0}. Considering the $p$ values of each measurement, we can reject $Ht_0$ in all three tasks. Obviously, corpora containing texts of tweets, movie or phone reviews perform better than corpora of generic texts on each corresponding task. The topic effect is small on sentiment analysis of tweets but much stronger on movie and phone reviews. 
\par
As we conclude this section, answering \hyperlink{rq2}{RQ2} we can say that pure word embedding features perform comparably well on sentiment analysis of tweets, song lyrics, and movie or phone reviews. We saw that top-ranking popular models like those of Common Crawl corpus perform very well on all the tasks. Also, our \emph{Item} collection of movie and product reviews works well. The collection of song lyrics (\emph{Lyric}) on the other hand produces worst results on each task, including analysis of lyrics. With respect to the training method, song lyrics are particularly sensitive, with Glove outrunning Word2vec in every experiment. We observed the same effect in a moderate scale on tweets as well, at least when using big corpora (e.g., \emph{Misch} models). On the other hand, we did not see any significant effect of training method on movie or phone reviews. Regarding corpus size, there was a moderate effect on both lyrics and tweets. Bigger text corpora usually provide embeddings that scale better on these two domains. Contrary, on movie and phone reviews we once again did not see any significant effect. Regarding the thematic relevance of the training corpora, we did not see any influence on song lyrics. The influence on tweets is moderate, with \emph{Twitter} models performing better than the rest. On phone and movie reviews we witnessed a very strong influence. \emph{Item} models trained with text collections of product reviews are superior to any other model.      
\section{Improving Quality of Word Embeddings}
\label{improvingQuality} 
The primary motive for developing distributed word representations was the need to fight the curse of dimensionality and data sparsity problems of the classical bag-of-words representation. Word embedding training methods do however exhibit certain deficiencies that limit their performance in various applications like sentiment analysis and more. First of all, Glove and Word2vec rely on co-occurrence statistics and window-based context for training and no use of word meaning is performed. As each word is represented by a single and unique vector, they do not count for homonymy (identical spelling but different meanings) or polysemy (association of one word with two or more distinct meanings). Furthermore, in subjective and emotional texts (e.g., tweets or song lyrics) it is frequent to find words with opposite polarity appearing close to each other (same context window). For this reason, they are mapped into word vectors that are similar. This is problematic as the vectors lose their discriminative ability when used as features in text classifiers. Such problems have been reported in \cite{tang2014learning} and other relevant studies. In the following sections, we present various methods that have been used to improve semantic quality of embeddings or generate better embeddings in the first place. 
\subsection{Enhancing quality with lexicons}
\label{qualityLexicons}
Linguistic lexicons have been widely used in text classification tasks as a rich source of features. WordNet is one of the most popular in the literature, rich with synonymy, hypernymy, paraphrase and other types of relations between words. In November 2012 it contained 155,287 English words organized in 117,659 synsets (words of the same lexical category that are synonymous). Authors in \cite{liuexploring} investigate the possibility of enriching Glove embeddings with semantic information contained in hypernym-hyponym relations of WordNet. More exactly, they modify co-occurrence matrix of Glove by slightly increasing the co-occurrence count of the word $w\textsubscript{2}$ if $w\textsubscript{1}$ and $w\textsubscript{2}$ appear in the same context and are hyponym and hypernym according to WordNet. Their results indicate that this method improves word vectors on small text corpora where generated word vectors do not perform very well. 
\par
A similar attempt is presented at \cite{faruqui2014retrofitting} where authors adopt a graph-based technique for exploiting lexical relations to improve quality of already trained word vectors. Their method is based on a graph of lexically-related words (appearing in a lexicon) and reorganizes word vectors to be similar to those of related word types and also similar to their original representations. This post-process retrofitting approach is fast and can be used to improve word vectors obtained by any training method. Authors report improvements in various tasks such as synonym selection, syntactic relations, word similarity etc. 
\par 
We observed the effect of this method ourself in two corpora of Table~\ref{table:corporaCuts}, namely \emph{Lyrics\_500M} and \emph{Tweets\_500M}. We used the Word2vec models of these corpora and also applied the retrofitting method using WordNet synonyms. The modified word vectors were reordered and the vector numbers were rounded to 5 decimal digits (from 7 in original vectors). To observe any effect on the performance we measured 5-fold cross-validation accuracy and macro F1 score on sentiment classification task (as in Section~\ref{corpusCharacteristics}) and also performance on word analogy task (as in Section~\ref{linguisticRegularity}). Obtained results are presented in Table~\ref{table:retrofitingMethod}. 
As we can see there is no obvious change in sentiment classification of tweets, except for a slight decrease in word analogy task. However, there is a considerable improvement of more than 2\% on song lyrics sentiment classification accuracy and a very small improvement on macro F1 as well. To ensure that this result is significant and reproducible, we performed paired t-test analyses on the five samples of each \emph{Lyrics} model. Obtained $t = 3.23$ and $p = 0.012$ values do confirm that the improvement is statistically significant. In response to \hyperlink{rq3}{RQ3} we can thus say that post-processing methods that exploit semantic word relations of lexicons may be used to improve quality of word embeddings at a certain scale on some tasks or domains. 
\begin{table}[ht] 
	\caption{Retrofiting method on Tweets and Lyrics}  
	\centering      
	\begin{tabular}{l | c c c c c c c}  
		\toprule
		\textbf{Corpus Name} & \textbf{Training} & \textbf{Size} & \textbf{Voc} & \textbf{Dim} & 
		\textbf{CV} & \textbf{F1} & \textbf{Analogy} 	\\  [0.1ex]
		\midrule  
		Tweets\_500M\_w2v & W2v & 500M & 96055 & 300 & 0.793 & 0.785 & 45.05  \\ [0.3ex]
		Tweets\_500M\_ret &  W2v + Ret & 500M & 96055 & 300 & 0.792  & 0.787 & 44.93 \\ [0.3ex]
		Lyrics\_500M\_w2v & W2v & 500M & 29306 & 300 & \underline{0.699} & 0.701 & 2.78 \\ [0.3ex]
		Lyrics\_500M\_ret & W2v + Ret & 500M & 29306 & 300 & \underline{\textbf{0.720}} & 0.708 & 2.71 \\ [0.3ex]				
		\bottomrule 
	\end{tabular} 
	\label{table:retrofitingMethod}
\end{table}
\subsection{Creating better embedding methods}
\label{qualityMethods}
A different direction for obtaining better word embeddings is to improve training methods in the first place. Various studies try to exploit word meanings or semantic relations directly in the training process. One of the earliest attempts was proposed in \cite{reisinger:naacl10} where they introduce multi-prototype vector space model. Authors try to discriminate word sense by clustering contexts of words and then generate different vectors for each sense of word based on context cluster. They evaluate on semantic similarity and near-synonymy prediction and report significant improvements. 
Similarly in \cite{Huang:2012:IWR:2390524.2390645}, they introduce a language model based on a neural network architecture that considers both local and global context of words. When word meaning in the local context is still ambiguous, information of global context is used to help disambiguation. Authors report an improved representation of word semantics and use it in their multi-prototype version where they consider multiple word senses based on context. Based on their experiments they report that the multi-prototype version outruns not only their single-prototype version but also other language representation models (e.g., C\&W) on word similarity tasks. 
\par
An even more recent work is \cite{tang2014learning} where authors examine the usability of word embeddings for sentiment analysis of tweets. They argue that word embeddings in their pure form are not effective enough for Twitter sentiment analysis as they ignore sentiment aspect of texts and thus map contrasting emotional terms in close vectors. As a remedy, they propose and present SSWE (Sentiment-Specific Word  Embeddings) that maps text words to the sentiment polarity of the document they appear in. Their goal of SSWE is to distinguish words of different or opposite polarities appearing in same contexts. By integrating sentiment information of text into the loss function of the neural networks, the generated vectors of those words are placed in opposite ends of the space and thus serve as better discriminators when used in sentiment analysis tasks. Authors experiment on Twitter sentiment analysis task utilizing Semeval-2013 dataset and report a considerable performance gap between SSWE (F1 score 84.98\%) and pure embedding methods like C\&W or Word2vec (F1 score 76.31\%) by at least 8\%. The obvious drawback of SSWEs is that they require emotionally labeled texts that are terribly scarce compared to the huge unlabeled texts that can be easily crawled from the web. 
\par
We decided to try SSWE on our own for sentiment analysis of tweets and song lyrics. We utilized the labeled texts of the datasets we experimented with in the previous sections which count for 250,000 and 1.2 million tokens of tweets and lyrics respectively. Considering the small size of the corpora we used a context window of 5 tokens and generated 100-dimensional vectors. The rest of the details is presented Table~\ref{table:sswe}. As we can clearly see, cross-validation scores are lower than those of Section~\ref{corpusCharacteristics}, obviously because of the much smaller training corpora. Nevertheless, it is important to note that SSWE provides slightly better accuracy (about 1\% higher) and macro F1, especially on the classification of tweets. Statistical analysis results ($t = 2.68$ and $p = 0.028$) 
confirm that the difference between \emph{Tweets\_w2v} and \emph{Tweets\_sswe} is significant. 
As a response to \hyperlink{rq4}{RQ4}, we affirm that alternative or application specific (e.g., SSWE) word embedding generation methods can produce better results on certain tasks or domains when used on large text collections. 
\begin{table}[!t] 
	\centering      
	\caption{SSWE vs WE on Tweets and Lyrics}  
	\begin{tabular}
		{l | c c c c c c}  
		\toprule
		\textbf{Corpus Name} & \textbf{Training} & \textbf{Size} & \textbf{Voc} & \textbf{Dim} & 
		\textbf{CV} & \textbf{F1}
		\\ [0.1ex] 
		\midrule 
		Tweets\_w2v & W2v & 250,000 & 12875 & 100 & \underline{0.727} & 0.726  \\ [0.3ex] 
		Tweets\_sswe &  SSWE & 250,000 & 14576 & 100 & \underline{\textbf{0.738}} & 0.729  \\ [0.3ex] 
		Lyrics\_w2v & W2v & 1.2M & 16350 & 100 & 0.685 & 0.664 		\\ [0.3ex] 
		Lyrics\_sswe & SSWE & 1.2M & 17905 & 100 & 0.692 & 0.673 	\\ [0.3ex] 
		\bottomrule
	\end{tabular} 
	\label{table:sswe}
\end{table}
\section{Discussion}
\label{discussion}
In this paper, we examined the quality of word embeddings when used in word analogy tasks, as well as their performance on sentiment analysis of tweets, song lyrics, movie reviews, and phone reviews. We were particularly interested to observe the role of factors like training method, training corpus size and thematic relevance on quality of the produced word embeddings. We also exercised the semantic and syntactic quality of models trained from different corpora with word analogy questions. Observing recent literature we found various post-processing techniques that try to improve semantic quality of word embedding by means of lexicons or specific training techniques that are able to generate better word embeddings in the first place. 
Based on our results, highest performance on word analogy task is achieved by models of text corpora that are multi multi-thematic, big in size and rich in vocabulary. Moreover, Glove slightly outruns Word2vec when used on very big text corpora. The opposite is true when small corpora models are used.     
Regarding sentiment analysis tasks, we observed that top results are achieved using models like \emph{Common Crawl} that are very big in size and rich in vocabulary. Our collection of item reviews resulted very effective also, despite its relatively small size. Contrary, song lyrics that very restricted in thematics and vocabulary performed badly on every task.  
Regarding the role of training methods, we observed that Glove models perform much better than Word2vec ones on song lyrics and somehow better on tweets. No significant difference between these two methods was observed on phone or movie reviews. 
Size of training corpora had a similar effect. Its role was important on song lyrics and visible on tweets analyzed with big models. However, we did not notice any significant influence on movie and phone reviews. 
Thematic relevance had a very strong impact on movie and phone reviews and a moderate impact on tweets. No effect was observed on sentiment analysis of lyrics, though. 
Various post-processing techniques that try to increase the semantic quality of word embeddings are proposed in the literature. We tried one of them that is based on \dq{injecting} intelligence of lexicons inside word vectors and saw that it indeed improves performance on song lyrics. 
Among alternative training methods, we explored SSWE which is designed to work specifically on sentiment analysis of tweets. Our results confirm that it indeed produces embeddings of higher quality when trained on large text bundles. 
Considering the above observations, we suggest that word embedding models should contain Wikipedia texts as part of them. These rich texts will certainly improve the performance of the model in semantic and syntactic analogy tasks. Furthermore, particular consideration should be given to the size of the training corpus, especially when analyzing tweets or song lyrics. In such tasks, Glove should be used over Word2vec. Other newer training method we did not explore here might also be considered. Contrary, when analyzing movie or product reviews, it is much more important to train review texts rather than bigger text bundles of other contents.

\section*{Acknowledgments} 
This work was supported by a fellowship from TIM (\url{https://www.tim.it/}). Part of computational resources was provided by HPC@POLITO (\url{http://hpc.polito.it}), a project of Academic Computing within the Department of Control and Computer Engineering at Politecnico di Torino. 

\section*{References}

\bibliography{wemb2}

\end{document}